%% file: main_arxiv.tex
\documentclass[10pt,a4paper]{article}

\usepackage{authblk}
\usepackage[margin=1.0in]{geometry}

\usepackage{amsmath}
\usepackage{amsfonts} 
\usepackage{amssymb}

\usepackage{bigints}
\usepackage{hyperref}
\usepackage{mathtools}
\usepackage[capitalise]{cleveref}
\usepackage[square,sort,comma,numbers,compress]{natbib}
\usepackage[utf8]{inputenc} 
\usepackage[T1]{fontenc}    
\usepackage{hyperref}       
\usepackage{url}            
\usepackage{booktabs}       
\usepackage{amsmath}
\usepackage{amsfonts}       
\usepackage{nicefrac}       
\usepackage{microtype}      
\usepackage{xcolor}         
\usepackage{bm}
\usepackage{algorithm}
\usepackage{algpseudocode}
\usepackage{wrapfig}
\usepackage{comment}
\usepackage{graphicx}

\newcommand{\app}[1]{SI Appendix \ref{#1}}
\labelformat{subsection}{\thesection#1}

\begin{document}

\title{Random Matrix Theory-guided sparse PCA for single-cell RNA-seq data}

\setcounter{Maxaffil}{1}
\author{Victor Chard\`es\thanks{vchardes@flatironinstitute.org}}
\affil[]{\small Center for Computational Biology, Flatiron Institute, New York, NY, USA, 10010}
\date{}

\maketitle

\input{src/abstract.tex}
\input{src/main.tex}

\bibliographystyle{pnas}
\bibliography{biblio}

\appendix
\newpage

\renewcommand{\thefigure}{S\arabic{figure}}
\renewcommand{\theHfigure}{S\arabic{figure}}
\setcounter{figure}{0}
\renewcommand{\theequation}{S\arabic{equation}}
\setcounter{equation}{0}
\renewcommand{\thealgorithm}{S\arabic{algorithm}}
\setcounter{algorithm}{0}

\input{src/si_text.tex}

\input{src/si_figures.tex}

\end{document}

%% file: src/abstract.tex
\begin{abstract}
Single-cell RNA-seq provides detailed molecular snapshots of individual cells but is notoriously noisy. Variability stems from biological differences and technical factors, such as amplification bias and limited RNA capture efficiency, making it challenging to adapt computational pipelines to heterogeneous datasets or evolving technologies. As a result, most studies still rely on principal component analysis (PCA) for dimensionality reduction, valued for its interpretability and robustness, in spite of its known bias in high dimensions. Here, we improve upon PCA with a Random Matrix Theory (RMT)-based approach that guides the inference of sparse principal components using existing sparse PCA algorithms. We first introduce a novel biwhitening algorithm which self-consistently estimates the magnitude of transcriptomic noise affecting each gene in individual cells, without assuming a specific noise distribution. This enables the use of an RMT-based criterion to automatically select the sparsity level, rendering sparse PCA nearly parameter-free. Our mathematically grounded approach retains the interpretability of PCA while enabling robust, hands-off inference of sparse principal components. Across seven single-cell RNA-seq technologies and four sparse PCA algorithms, we show that this method systematically improves the reconstruction of the principal subspace and consistently outperforms PCA-, autoencoder-, and diffusion-based methods in cell-type classification tasks.
\end{abstract}

%% file: src/main.tex
\section{Introduction}
Single-cell RNA-seq measures the number of mRNA transcripts per gene in individual cells extracted from a tissue. Given a matrix $X \in \mathbb{R}^{n \times p}$ of measurements across $n$ cells and $p$ genes, a central goal is to classify cells into cell types and identify marker genes by differential expression. In practice, this is usually done in an unsupervised manner: the data is projected onto a lower-dimensional space using dimensionality reduction methods, most commonly PCA, followed by clustering \cite{luecken2019current, andrews2021tutorial}. PCA owes its success in this setting to its interpretability and robustness. It is a linear method that amounts to computing the leading eigenvectors of the sample covariance matrix
\begin{equation}
S_{kq} = \frac{1}{n-1} \sum_{i=1}^{n} (X_{ik} - \bar X_k)(X_{iq} - \bar X_q),
\label{cent_cov}
\end{equation}
where $\bar X \in \mathbb{R}^p$ is the sample mean expression. This decomposition identifies orthogonal directions, or principal components (PCs), that capture the largest variation in the data, enabling projection of cells onto fewer axes that retain the main patterns of variability. This approach is the primary dimensionality reduction method used in popular single-cell RNA-seq analysis packages such as Scanpy \cite{wolf2018scanpy} and Seurat \cite{stuart2019comprehensiveb}.

When many more cells are sampled than genes are measured, $S$ converges to $\mathbb{E}[S]$, and its eigenspectrum and eigenvectors reliably estimate those of $\mathbb{E}[S]$. In typical single-cell RNA-seq experiments, however, the number of cells is comparable to the number of genes. In this high-dimensional regime, the leading principal components of $S$ are poor estimators of those of $\mathbb{E}[S]$. In practice, the extent of this error is quantified by computing the overlap between the two principal subspaces. This overlap is maximal when $p/n \rightarrow 0$ and decreases toward zero as $p/n$ increases. Because single-cell RNA-seq is destructive, the same cells cannot be repeatedly measured to average out this variability, leaving accurate estimation of the principal components of $\mathbb{E}[S]$ a central challenge. 

For this reason, the question we address is: how can we accurately estimate the PCs of $\mathbb{E}[S]$ when the number of cells $n$ and genes $p$ are large but comparable? This high-dimensional regime falls within the scope of Random Matrix Theory (RMT), which provides tools to predict the eigenspectrum of $S$, as well as the overlap between the leading eigenspaces of $S$ and $\mathbb{E}[S]$. A substantial body of work leverages these results to construct better estimators of $\mathbb{E}[S]$, with most approaches correcting leading eigenvalues of $S$ so as to minimize a chosen distance to $\mathbb{E}[S]$ \cite{bun2017cleaning, ledoit2015spectrum, nadakuditi2014optshrink, gavish2014optimal, gavish2017optimal, leeb2021optimal, bun2016rotational}. Importantly, these improved estimators are rotationally invariant, meaning that the eigenvectors of $S$ remain uncorrected. In the specific context of single-cell RNA-seq, several studies have used RMT either to construct such estimators \cite{stanley2025principled} or as a tool to distinguish noise from signal \cite{mircea2022phiclust, kim2024sclens, aparicio2020random, leviyang2024analysis}.

In this paper, we take a different approach and focus on directly denoising the principal components of $S$. This problem has also been studied extensively through sparse PCA methods, which seek sparse principal components of $S$. To do so, these approaches augment PCA with sparsity constraints imposed on the principal components. Many sparse PCA algorithms have been proposed \cite{journee2010generalized, chavent2021optimala, jolliffe2003modified, zou2006sparse, chen2020alternating, jenatton2010structured}, each implementing a variation of the same idea. However, none has been systematically applied to realistic single-cell RNA-seq datasets. A likely reason is that sparse PCA is highly sensitive to the choice of the penalty parameter: overestimating it can introduce misleading artifacts that may be mistaken for biological signal. We resolve this issue by showing, on a variety of single-cell RNA-seq datasets, that RMT enables robust, hands-off inference of sparse principal components that better approximate the leading eigenspace of $\mathbb{E}[S]$ than standard PCs. Through systematic benchmarks on datasets with ground-truth cell type annotations, we find that sparse PCA consistently outperforms autoencoder-, diffusion-, and PCA-based methods on the task of cell type annotation.

\section{Results}
\subsection{Assumptions and methodology}
We assume that each cell follows the same stochastic gene regulatory process, so that the pattern of gene–gene correlations does not vary across cells. Implicitly, this corresponds to a separable covariance structure \cite{paul2009no}, i.e., $\mathbb{E} [(X_{ij} - \mathbb{E}[X_{ij}])(X_{kl} - \mathbb{E}[X_{kl}])] = A_{ik} B_{jl}$, where $A$ is the cell-cell covariance matrix and $B$ the gene-gene covariance matrix. Under this assumption, the data matrix $X$ can be written as
\begin{equation}
X = A^{1/2} Y B^{1/2} + P, \label{sep_X}
\end{equation}
where $Y_{ij}$ are i.i.d. random variables with zero mean and unit variance, and $P = \mathbb{E}[X]$ is a low-rank matrix. The signal, also assumed low-rank, may lie in the mean $P$ (the information-plus-noise model \cite{benaych-georges2012singular}), in the covariance $B$ (the spiked separable covariance model \cite{johnstone2001distributiona, bai2012sample, yao2015largea}), or in both. While this central assumption is largely supported by recent findings on various single-cell RNA-seq datasets \cite{landa2022biwhitening, stanley2025principled}, we will also confirm its validity on the data used in this paper. We adopt standard assumptions for the separable covariance matrice model \cite{couillet2015analysis, ding2021spiked}. Notably: the fourth moments of the entries of $Y$ are bounded, the spectral distributions of $A$ and $B$ converge to compactly supported probability distributions as $n \to \infty$ with $q = p/n$ fixed, and these limits are associated with densities $\rho_A$ and $\rho_B$, respectively. Under these assumptions, the spectral distribution of $S$ converges to a compactly supported probability distribution with density $\rho_S$ \cite{couillet2015analysis}.

A central goal of RMT is to separate the contributions to the eigenspectrum of $S$ arising from the noise and from the low-rank signal. Under the assumptions above, RMT provides an analytical mapping between signal eigenvalues and the eigenvalues $\lambda$ of $S$ that lie outside the support of $\rho_S$ \cite{yao2015largea, ding2021spiked}. The eigenvectors associated with these outlier eigenvalues span the outlier eigenspace, which we aim to denoise with sparse PCA. Crucially, RMT not only predicts the mapping between signal and outlier eigenvalues, but also the angle between each signal eigenvector and the outlier eigenspace, and vice versa \cite{bloemendal2016principal, ding2021spiked}. This suggests that we can search for sparse PCs consistent with these angle predictions, following a maximum a posteriori inference approach \cite{monasson2015estimating}. In practice, however, this task is computationally demanding. We simplify it by selecting the sparsity level in sparse PCA so that the inferred subspace and the outlier subspace approximately match the angle predicted by RMT.

However, two obstacles arise: (i) we need estimators of $\rho_A$ and $\rho_B$ to compute $\rho_S$ and identify the outlier eigenspace, and (ii) the RMT mapping between signal and outlier eigenspaces depends on whether the signal lies in $P$, in $B$, or in both, as detailed in \app{sec:limits}. To address the first obstacle, building on recent advances in matrix biwhitening \cite{stanley2025principled}, we develop a novel algorithm to jointly estimate $A$ and $B$ without requiring assumptions on the noise distribution. However, we show that empirical estimators of $\rho_A$ and $\rho_B$ derived from these estimates fail to reconstruct the support of $\rho_S$, a crucial step to identify the outlier eigenspace. Lacking better estimators, we instead use our estimates of $A$ and $B$ to form the biwhitened matrix $X_{\mathrm{bw}} = A^{-1/2} X B^{-1/2}$, for which $\rho_S$ is known analytically, as detailed in Section~\ref{sec:methods}. This ensures reliable estimation of the outlier eigenspace for the biwhitened matrix.

Moreover, this biwhitening step also resolves the second obstacle: recent results show that for left-whitened data $X_{\mathrm{lw}} = A^{-1/2}X$, there exists a unique mapping between the signal and outlier eigenspaces, irrespective of whether the signal lies in $P$, $B$, or both \cite{liu2023asymptotica}. This result carries over to biwhitened data $X_{\mathrm{bw}}$, meaning that by working with $X_{\mathrm{bw}}$, we can guide a sparse PCA algorithm without assuming a specific model for where the signal lies. Further discussion and illustrations of these findings are provided in \app{sec:limits}. For these reasons, we propose a two-step approach: (i) estimate $A$ and $B$ using our novel biwhitening algorithm, and (ii) use RMT results to guide the choice of the sparsity parameter in sparse PCA and denoise the PCs of the biwhitened matrix $X_{\mathrm{bw}}$.

\begin{figure*}
\begin{minipage}{\linewidth}
\begin{algorithm}[H]
\caption{Sinkhorn-Knopp Biwhitening}\label{algo_sknopp}
\begin{algorithmic}[0]
\State \textbf{input:} $n \times p$ matrix $X$
\State \textbf{output:} scaling vectors $c, d$

\State Square root, inverse and power operations on column vectors are performed entry-wise
\State The operator $\mathrm{diag}(v)$ denotes the diagonal matrix with diagonal entries $v$
\State $U_{ij} \gets X_{ij}^2$, for all $i = 1,\dots,n$ and $j = 1,\dots,p$
\State $c_i^{(0)} \gets 1,\; d_j^{(0)} \gets 1$, for all $i = 1,\dots,n$ and $j = 1,\dots,p$

\While{stopping criterion not reached}
    \State $d^{(k+1)} \gets \left[n \left(1 + \big(\mathrm{diag}(d^{(k)}) X^T c^{(k)}\big)^2/n^2\right) \big/ \left(U^T (c^{(k)})^2\right)\right]^{1/2}$
    \State $c^{(k+1)} \gets \left[p \left(1 + \big(\mathrm{diag}(c^{(k)}) X d^{(k+1)}\big)^2/p^2\right) \big/ \left(U (d^{(k+1)})^2\right)\right]^{1/2}$
    \State $k \gets k+1$
\EndWhile
\State $z \gets \mathrm{diag}(c^{(k)})\, X \,\mathrm{diag}(d^{(k)})$
\State $\lambda_{\mathrm{med}} \gets$ median of the unit variance Marchenko–Pastur distribution with $q=p/n$ if $p \leq n$, else $1/q$
\State $\ell_{\mathrm{med}} \gets$ median eigenvalue of $Z^T Z / n$ if $p \leq n$, else of $Z Z^T / p$
\State $\sigma^2 \gets \ell_{\mathrm{med}}/\lambda_{\mathrm{med}}$
\State \Return $c^{(k)}/\sigma,\; d^{(k)}$
\end{algorithmic}
\end{algorithm}
\end{minipage}
\end{figure*}

\subsection{A novel biwhitening algorithm ensures a robust separation of signal and noise}
We assume that $A$ and $B$ are diagonal matrices with strictly positive entries. To estimate them, we simultaneously optimize for two diagonal matrices $C$ and $D$ with positive entries such that the cell-wise and gene-wise variances of $Z = C X D$ are approximately one, i.e.,
\begin{gather}
\frac{1}{n} \sum_{i=1}^{n} \left(Z_{ij} - \frac{1}{n} \sum_{k=1}^{n} Z_{kj}\right)^2 \simeq 1, \forall j,\\
\frac{1}{p} \sum_{j=1}^{p} \left(Z_{ij} - \frac{1}{p} \sum_{k=1}^{p} Z_{ik}\right)^2 \simeq 1, \forall i.
\end{gather}
This procedure is known as biwhitening \cite{landa2022biwhitening, stanley2025principled}. Our first contribution is to reformulate the problem in terms of the diagonal entries $c_i$ ($1 \leq i \leq n$) and $d_j$ ($1 \leq j \leq p$) of $C$ and $D$, respectively:
\begin{gather}
\frac{1}{n} \sum_{i=1}^{n} c_i^2 X_{ij}^2 d_j^2 \simeq 1 + \left(\frac{1}{n} \sum_{i=1}^{n} c_i X_{ij} d_j\right)^2, \forall j, \\
\frac{1}{p} \sum_{j=1}^{p} c_i^2 X_{ij}^2 d_j^2 \simeq 1 + \left(\frac{1}{p} \sum_{j=1}^{p} c_i X_{ij} d_j\right)^2, \forall i.
\end{gather}
In this form, solving for $C$ and $D$ is equivalent to a bi-proportional scaling problem on the entry-wise squared data matrix \cite{bacharach1965estimating}, with scaling matrices $C^2$ and $D^2$, but with moving targets for the row and column sums. Exploiting this similarity, we adapt the Sinkhorn–Knopp algorithm to handle these moving targets \cite{sinkhorn1967concerning}. The resulting procedure, detailed in Alg.~\ref{algo_sknopp}, returns diagonal matrices $C$ and $D$ such that $C^{-2} \simeq A$ and $D^{-2} \simeq B$. Finally, because the overall noise variance after biwhitening is close to, but not exactly, one, we further normalize $C$ by dividing it by a robust estimator of the standard deviation $\sigma$ of the data \cite{gavish2014optimal, gavish2017optimal}.

\begin{figure*}[t!]
  \includegraphics[width=\textwidth]{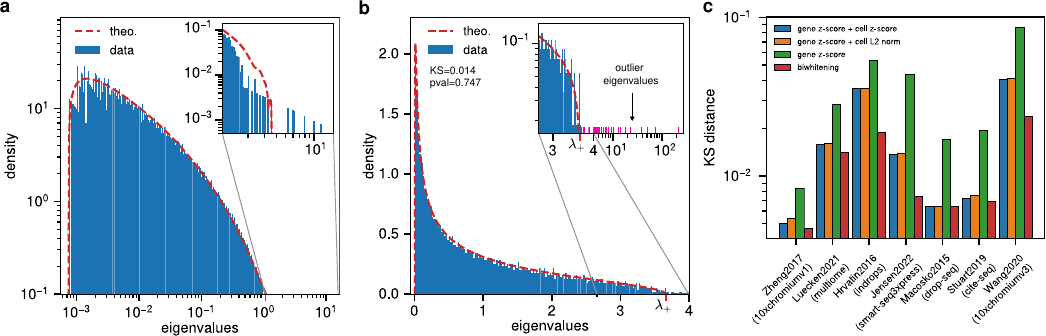}
  \caption{\textbf{Validity of the separable covariance model.} The factors $A$ and $B$ are estimated using our Sinkhorn-Knopp Biwhitening algorithm. {\bf a.} Numerical solution for the density $\rho_S$ from Eq.~\ref{sol_stieljes} and Eq.~\ref{fpoint_g2} using empirical estimators for $\rho_A$ and $\rho_B$. {\bf b.} Eigenspectrum of the covariance matrix after biwhitening, shown alongside the Marchenko-Pastur distribution (dashed red line). The inset highlights the spectral edge, with outlier eigenvalues in pink. The close agreement with the Marchenko-Pastur prediction is supported by the small Kolmogorov-Smirnov (KS) distance and large $p$-value \cite{luecken2021sandbox}. {\bf c.} KS distance between the covariance eigenspectrum after biwhitening (red), gene-wise $z$-scoring (red), gene-wise $z$-scoring followed by cell-wise $z$-scoring (blue), and gene-wise $z$-scoring followed by cell-wise $L_2$ normalization (orange). All datasets were processed using $2500$ highly variable genes, following the flow charts in Fig.~\ref{fig:figSpipe} (see \app{sec:reproducibility}) For {\bf a} and {\bf b}, the dataset used is Luecken2021.}
  \label{fig:fig1}
\end{figure*}

Having estimated $A$ and $B$, we can construct empirical estimators for $\rho_A$ and $\rho_B$, respectively $\rho_A(t) \approx \sum_i^n \delta(t - c_i^{-2})/n$ and $\rho_B(t) \approx \sum_j^p \delta(t - d_j^{-2})/p$, and use them to solve numerically for $\rho_S$, as detailed in Section.~\ref{sec:methods}. In Fig.~\ref{fig:fig1}a, we overlay this theoretical prediction with the eigenspectrum of $S$ from a single-cell RNA-seq dataset after library size normalization and $\log(x+1)$ transform. The RMT solution closely matches the data, confirming that the biwhitening procedure provides satisfying estimates of $A$ and $B$. However, as shown in the inset of Fig.~\ref{fig:fig1}a and noted in the introduction, the RMT solution fails to correctly estimate the support of $\rho_S$. This failure arises because the empirical estimators of $\rho_A$ and $\rho_B$ formed from $C$ and $D$ contain outlier eigenvalues, which are absent from the true limiting densities as $n \to \infty$ with $q=p/n$, thereby preventing the identification of the outlier eigenspace.

For this reason, we work with biwhitened data, obtained using the matrices $C$ and $D$: $X_{\mathrm{bw}} = C X D$. For such data, we expect the eigenspectrum of $S$ to closely follow the analytically known Marchenko–Pastur distribution. This is illustrated in Fig.~\ref{fig:fig1}b, with the inset showing a zoom on the rightmost edge of the spectrum. The Marchenko–Pastur law accurately captures the bulk support of the eigenvalues of $S$, leaving only a few eigenvalues above the spectral edge $\lambda_{+} = (1 + \sqrt{q})^2$. Since this support is known analytically, we can directly identify the outlier eigenvalues $O$ as those exceeding $\lambda_{+}$. Their associated eigenvectors span the outlier eigenspace, equivalently represented by its orthogonal projector $W$. Finally, the agreement between the Marchenko–Pastur distribution and the observed eigenspectrum, quantified by the Kolmogorov–Smirnov (KS) distance and its associated p-value in the inset, confirms that the separable covariance model is a statistically valid assumption for single-cell RNA-seq data. From a modeling perspective, this means that, with the number of genes and cells used in this example, correlation structure beyond a diagonal and separable covariance model is not statistically significant. Importantly, this does not rule out the existence of such correlations, which could be revealed by analyzing more cells than in Fig.~\ref{fig:fig1}.

To illustrate the robustness of our approach, we applied it to seven datasets corresponding to seven different single-cell RNA-seq technologies \cite{hagemann-jensen2022scalable, zheng2017massively, hrvatin2018singlecella, luecken2021sandbox, macosko2015highly, stuart2019comprehensiveb, wang2020singlecellb} (see Section~\ref{sec:methods}). In Fig.~\ref{fig:fig1}c, we compare the performance of our biwhitening algorithm against several alternative whitening strategies: (i) gene-wise $z$-scoring, (ii) gene-wise $z$-scoring followed by cell-wise $z$-scoring \cite{mircea2022phiclust}, and (iii) gene-wise $z$-scoring followed by cell-wise $L_2$ normalization \cite{kim2024sclens}. Method (i) is the standard preprocessing step before PCA with single-cell RNA-seq data. Method (ii) was proposed in \cite{mircea2022phiclust} to better align the eigenspectrum with the Marchenko–Pastur distribution compared to (i), which we confirm in Fig.~\ref{fig:fig1}c. Method (iii), recently introduced as a new normalization approach \cite{kim2024sclens}, is in fact mathematically very similar to (ii) and produces identical performance in terms of KS distance. Across all datasets, our biwhitening approach consistently outperforms these whitening variants, with the largest gains observed when gene-wise $z$-scoring alone fails to sufficiently whiten the data, as reflected by larger KS distances.

Our approach closely resembles the recently introduced BiPCA algorithm \cite{stanley2025principled}, which also uses the Sinkhorn-Knopp algorithm to estimate biwhitening factors $C$ and $D$. However, BiPCA can only operate on data for which the variance of gene expression is quadratically related to its mean. This assumption holds at the level of counts, but breaks down at later stages of data processing. Instead, our method self-consistently estimates the variance of gene expression without assuming any specific relationship with the mean expression. This modification is the main novelty of our approach, as it allows us to biwhiten the data at any stage of preprocessing: directly on counts, after library-size correction, or even after log-normalization, as shown in Fig.~\ref{fig:fig1}. In Fig.~\ref{fig:figScounts}a, we show that our biwhitening algorithm performs almost identically to BiPCA when applied to count data. In particular, Fig.~\ref{fig:figScounts}a,b demonstrate that it recovers biwhitening factors $C$ and $D$ that are nearly identical to those of BiPCA, irrespective of the number of highly variable genes selected.

\subsection{RMT-guided sparse PCA estimates the signal eigenspace better than standard PCA}
Having estimated $A$ and $B$, we now use them to denoise the biwhitened data $X_{\mathrm{bw}}$. Recall that our strategy to guide sparse PCA with RMT is as follows: apply any sparse PCA method to $X_{\mathrm{bw}}$ and select its sparsity parameter $\gamma$ so that the inferred subspace $\hat{Q}$ forms the angle predicted by RMT with the outlier eigenspace $W$. Specifically, the mapping between outlier and signal eigenvalues states that each outlier eigenvalue $\lambda$ corresponds to a unique signal eigenvalue $\alpha$ satisfying $\alpha = -1/\underline{m}(\lambda)$, where $\underline{m}(\lambda)$ is the complementary Stieltjes transform of $\rho_S$ (see Section~\ref{sec:methods}). Likewise, each outlier eigenvector is expected to have squared overlap $\|Qv\|^2_2 = \alpha \psi'(\alpha)/\psi(\alpha)$ where $\psi(\alpha)$ is the functional inverse of $-1/\underline{m}(\lambda)$, and is known analytically (see Section~\ref{sec:methods}). Based on this, we propose choosing $\gamma$ such that the following relation holds:
\begin{align}
\mathrm{tr}(\hat{Q}W) \gtrsim \mathrm{tr}(QW) &= \sum_{\underset{\alpha = -1/\underline{m}(\lambda)}{\lambda \in O}} \alpha \frac{\psi'(\alpha)}{\psi(\alpha)} \label{criterion_gen} \\
&=\sum_{\underset{\alpha = -1/\underline{m}(\lambda)}{\lambda \in O}} \frac{(\alpha -1)^2 - q}{(\alpha - 1)(\alpha - 1 + q)}, \label{criterion_bw}
\end{align}
where $q = p/n$. If sparse PCA could exactly recover $Q$, this condition would hold with equality. In practice, exact recovery is not possible, so $\gamma$ must be chosen such that $\mathrm{tr}(\hat{Q}W)$ remains close to this lower bound. As an important side note, unlike Eq.~\ref{criterion_bw} which uses the analytical expression for $\psi(\alpha)$ for biwhitened data, Eq.~\ref{criterion_gen} is fully general and thus also applies when working with left-whitened data, $X_{\mathrm{lw}} = A^{-1/2} X$, instead of biwhitened data. This opens the possibility of applying the criterion directly at the level of counts, provided a better estimator for the support of $\rho_S$ allows the identification of signal from noise with left-whitened data.

Thanks to its generality, this criterion can be applied to any sparse PCA algorithms for which the sparsity level is controlled with a single parameter $\gamma$. The methods we consider are: i) max-variance (Gpower) \cite{journee2010generalized, chavent2021optimala}, ii) dictionary-learning (sklearn) \cite{jenatton2010structured}, iii) regression-based (AManPG) \cite{zou2006sparse, chen2020alternating}. We also developed a naive approach based on the FISTA algorithm, detailed in Alg.~\ref{fista_spca} to solve a maximum-variance formulation of sparse PCA inspired by the SCoTLASS algorithm \cite{jolliffe2003modified, beck2009fasta}. When un-penalized, all these algorithms solve a different, yet mathematically equivalent, formulation of PCA \cite{chavent2021optimala}. When penalizing with an $L_1$ norm the loading vectors, they each solve nonequivalent problems, and we do not expect them to provide the same solution.

\begin{figure*}[t!]
  \includegraphics[width=\textwidth]{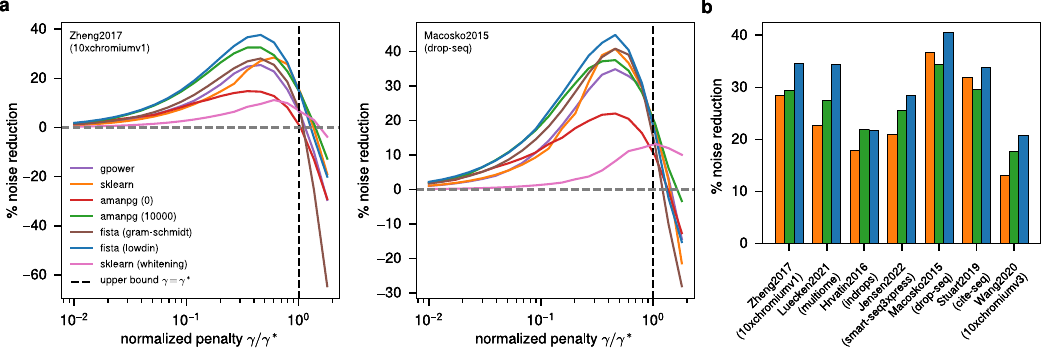}
  \caption{\textbf{Noise reduction obtained with sparse PCA.} {\bf a.} Noise reduction as a function of $\gamma/\gamma^*$ for five different sparse PCA algorithms. For AManPG we use two $L_2$ penalties in its elastic-net regularization: no penalty (red) and a penalty of $10^4$ (green). For our FISTA implementation we compare two orthogonalization methods: Gram–Schmidt (brown) and Löwdin (blue). In pink, we apply sklearn algorithm after gene-wise z-scoring (whitening) instead of biwhitening. {\bf b.} Error reduction at $\gamma = 0.6 \gamma^*$ for all datasets using the top-performing methods, with an average reduction of $\sim 30\%$. The colors correspond to the legend in {\bf a}. For all datasets we used $2000$ highly variable genes, and preprocessing before biwhitening/whitening and sparse PCA followed the same steps as in Fig.~\ref{fig:fig1}.}
  \label{fig:fig2}
\end{figure*}

We now assess the viability of this criterion on realistic single-cell RNA-seq datasets, using the same seven technologies as before. For each dataset, we subsample $3000$ cells and select the $2000$ most highly variable features. After standard preprocessing: library-size correction and log-normalization followed by biwhitening, we apply our approach to each subsampled dataset. To isolate the effect of biwhitening, we also substitute the biwhitening step by gene-wise $z$-scoring before applying our RMT-guided sparse PCA approach. Since the original datasets contain $\geq 30000$ cells, we have access to three subspaces: (i) the inferred subspace $\hat{Q}$, (ii) the outlier eigenspace $W$, and (iii) the outlier eigenspace for the full dataset, $W_{\mathrm{full}}$. In the limit $n \to \infty$, the outlier eigenspace converges to the true signal eigenspace $Q$, so we use $W_{\mathrm{full}} \simeq Q$ as a proxy for the signal. With this insight, we evaluate the efficacy by measuring how much closer $\hat{Q}$ is to $W_{\mathrm{full}}$ compared to $W$. We define noise reduction as the relative improvement over standard PCA:
\begin{equation}
\text{noise reduction} = 1 - \frac{d^2(\hat{Q}, W_{\mathrm{full}})}{d^2(W, W_{\mathrm{full}})},
\end{equation}
where $d(\hat{Q},W)$is the chordal distance between subspaces of different dimensions \cite{ye2016schubert}. Any other valid distance could also be used.

In Fig.~\ref{fig:fig2}a, for the Zheng2017 and Macosko2015 datasets (see Section~\ref{sec:methods}) we report the noise reduction for each sparse PCA algorithm as a function of $\gamma / \gamma^*$, where $\gamma^*$ is the optimal sparsity parameter for which criterion Eq.~\ref{criterion_bw} is exactly satisfied. The curves collapse satisfactorily across methods, underscoring the generality of criterion Eq.~\ref{criterion_bw}. We also observe a sharp performance drop for penalty parameters above $\gamma^*$, illustrating the critical role of this criterion: applying sparse PCA with an overestimated penalty invariably destroys the biological signal. As noted earlier, because the inferred eigenspace does not perfectly match the true signal subspace, the theoretical criterion is not exact. In practice, selecting $\gamma \simeq 0.6 \gamma^*$ appears to work well across all algorithms, and we suggest adopting this empirical criterion when applying sparse PCA. Importantly, while this empirical criterion yields near-optimal performance, we also observe in Fig.~\ref{fig:fig2}a that any $\gamma \lesssim \gamma^*$ improves performance over PCA. Finally, Fig.~\ref{fig:figSscan} and Fig.~\ref{fig:figSscan_v3} show that these results extend to all seven datasets and remain robust to changes in the set of highly variable genes.

We also find that applying sparse PCA after gene-wise $z$-scoring leads to a dramatic loss in performance, highlighting the necessity of the biwhitening procedure. For the AManPG-based approach, the best results are obtained with elastic net regularization and a large $L_2$ penalty. As shown in Fig.~\ref{fig:figSscan_amanpg}, taking the $L_2$ penalty to diverge yields the strongest performance. By contrast, applying AManPG with purely $L_1$ regularization consistently underperforms relative to other sparse PCA methods. We interpret this as evidence that enforcing sparsity on regression weights, as in AManPG, is fundamentally different from imposing sparsity on loading vectors, even though the unpenalized problems are equivalent, a distinction recently emphasized in \cite{park2024critical}. Surprisingly, our naive FISTA implementation for sparse PCA, where principal components are orthogonalized at each iteration using Löwdin’s method \cite{lowdin1970nonorthogonality}, outperforms all other approaches. This unexpected success suggests that, despite its heuristic nature, this new sparse PCA method stands as a top competitor among existing approaches.

Overall, we obtain the best results with sklearn-, FISTA- and AManPG-based sparse PCA, with a clear preference for the pipeline combining library-size normalization, log-normalization, and biwhitening. For $\gamma \simeq 0.6 \gamma^*$, this setup achieves an average noise reduction of $30\%$ over all the datasets, as shown in Fig.~\ref{fig:fig2}b. With this benchmark, we showed that under the separable covariance model, sparse PCA improves the recovery of the low-rank signal over PCA. This, however, doesn't necessarily indicates that sparse PCA improves biological signal recovery. For this reason, we now turn to evaluate the performance of our approach on the task of cell type annotation.

\begin{figure*}[t!]
  \includegraphics[width=\textwidth]{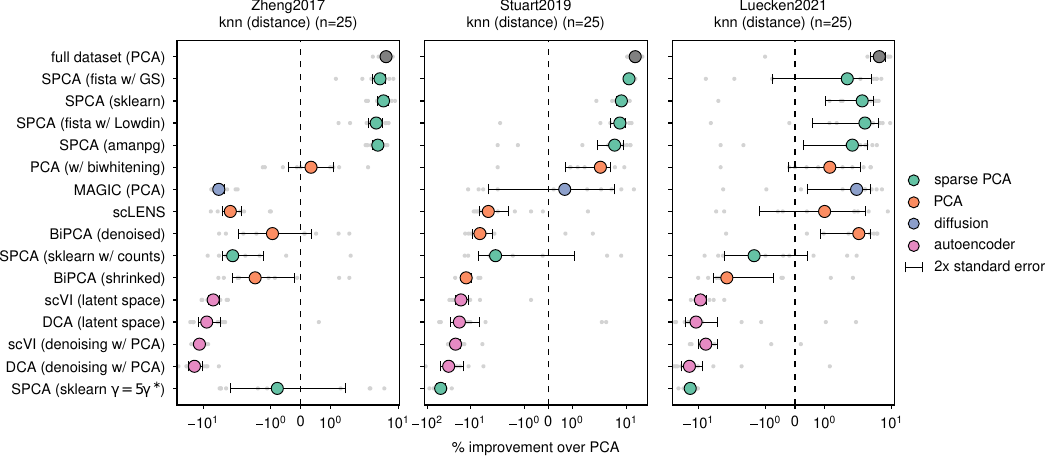}
  \caption{\textbf{Comparison of cell type annotation performance for state-of-the-art methods.} Improvement over PCA in out-of-bag error for bagging of $30$ $k$-NN classifiers with network weights inversely proportional to the distance between points and $n=25$ neighbors. The methods are ranked from best (top) to worst (bottom), as measured across classifiers using constant weights, inverse-distance weights, UMAP weights, and $n=10, 25, 45$ neighbors. To further reduce the variance of the classification error, we exclude from the analysis cell types represented by fewer than $30$ cells. The best-performing method (grey) uses all cells in the dataset ($\geq 30000$) to compute PCs, which are then used to project the subset of cells used in the benchmark. The next best performers are all RMT-guided sparse PCA methods. We use $3000$ cells and $2000$ highly variable genes. Complete benchmark flow charts for each dataset are shown in Fig.~\ref{fig:figSpipe_zheng}-\ref{fig:figSpipe_stuart}. Each experiment was repeated $10$ times with a different cell subset selected at random. Benchmark results for other types of $k$-NN classifiers are shown in Fig.~\ref{fig:figSknn}.}
  \label{fig:fig3}
\end{figure*}

\subsection{RMT-guided sparse PCA outperforms state-of-the-art methods on cell type annotation}
We use three datasets with ground-truth cell type labels: (i) Zheng2017, human Peripheral Blood Mononuclear Cells (PBMC) annotated by correlation-based assignment to purified cell types \cite{zheng2017massively}; (ii) Stuart2019, Human Bone Marrow Mononuclear Cells (BMMC) CITE-seq data, annotated with an unsupervised approach integrating protein and RNA modalities \cite{stuart2019comprehensiveb} ; and (iii) Luecken2021, Human BMMC Multiome data, annotated using an unsupervised method with cross-modality validation \cite{luecken2021sandbox}. On this task of cell type annotation, we compare our approach, biwhitening followed by sparse PCA (termed SPCA), with several methods designed to denoise single-cell RNA-seq data: (i) autoencoder-based methods: scVI \cite{lopez2018deep} and DCA \cite{eraslan2019singlecella}, (ii) the diffusion-based method MAGIC \cite{dijk2018recovering}, and (iii) PCA- and RMT-based methods: scLENS \cite{kim2024sclens} and BiPCA \cite{stanley2025principled}.

For all methods, after selecting $2000$ highly variable genes, we apply the denoising algorithm either to raw count data (scVI and DCA) or to library-size–corrected and log-normalized data (MAGIC). As a control for the effect of biwhitening alone, after library-size correction and log-normalization, we also apply biwhitening followed by PCA. Performance is measured relative to the canonical pipeline, simply referred to as PCA: library-size correction, log-normalization, and gene-wise $z$-scoring followed by PCA, using the same number of components as in the other methods. Flow charts describing the benchmark design and processing pipelines for each dataset provided in Fig.~\ref{fig:figSpipe_zheng}-\ref{fig:figSpipe_stuart}.

Because cell type annotation often relies on community clustering with a $k$-Nearest-Neighbor ($k$-NN) graph, it is especially relevant to evaluate performance at the level of nearest-neighbor relationships \cite{ahlmann-eltze2023comparison}. We assess the quality of cell type classification by measuring the accuracy of $k$-NN classifiers trained on the ground-truth labels. Following \cite{linderman2022zeropreservinga}, classification performance is quantified using Bootstrap Aggregation, with the out-of-bag error serving as the measure of accuracy \cite{breiman1996bagging}. In Fig.~\ref{fig:fig3}, we report the relative change in out-of-bag error for each method compared to PCA. By this metric, classification with sparse PCs performs on par with classification using PCs computed from the full dataset via $W_{\mathrm{full}}$. The latter serves as a control, representing the best performance achievable with the same number of PCs but many more cells. For a fair comparison, we performed hyperparameter searches over 50 models for both DCA and scVI, using their built-in parameter tuning tools. Since autoencoders embed data into a lower-dimensional space distinct from the PCs, we evaluated cell type classification both on PCs derived from denoised data and directly in their intrinsic latent space. 

Regardless of the latent space considered, all autoencoder-based methods underperformed relative to PCA. Similarly, MAGIC also underperformed compared to PCA for this task. These results are consistent with those in Fig.~\ref{fig:figSsilhouette}, where performance is assessed by measuring the average silhouette score of the ground-truth clustering on each low-dimensional embedding.We also notice that biwhitening followed by PCA leads to better performance than PCA alone, which we interpret as improved variance stabilization and better identification of the signal eigenspace. Importantly, we observe an overall underperformance of methods applied directly to counts (BiPCA and SPCA on counts) compared to PCA applied to log-normalized data. Put simply, this suggests that log-normalization improves the quality of PCA-derived low-dimensional embeddings. This is consistent with the observation that log-normalization followed by PCA still outperforms alternative methods \cite{ahlmann-eltze2023comparison}. Finally, we show that SPCA with an overestimated sparsity parameter, $\gamma = 5\gamma^*$, leads to drastic underperformance, highlighting that tuning $\gamma \lesssim \gamma^*$ is critical to SPCA's success. Because scLENS and BiPCA are designed to operate on more than the $2000$ highly variable genes typically used, Fig.~\ref{fig:figSknn_10000} verifies that our conclusions are robust to the number of genes included, and Fig.~\ref{fig:figSknn_v3} confirms that our results do not depend on the method used to select highly variable genes.

Our results support the conclusion that our approach, biwhitening followed by RMT-guided sparse PCA, is more effective than autoencoder- and diffusion-based methods for cell type annotation. Unlike autoencoders, which require fitting probabilistic models with thousands of parameters, our method is \emph{almost} parameter-free. That said, in its current form it is tailored specifically to cell type classification, whereas autoencoders can address a broader range of tasks, including direct denoising at the level of counts.

\section{Discussion}
In this paper, we proposed an RMT-guided methodology for applying sparse PCA to single-cell RNA-seq data \cite{johnstone2018pca}. Rather than constructing rotationally invariant estimators of the covariance matrix, we focused on denoising its leading eigenvectors. Our first contribution is a novel biwhitening algorithm, inspired by Sinkhorn–Knopp biproportional scaling, that stabilizes variance across cells and genes. Using this algorithm, we showed that single-cell RNA-seq data is consistent with a separable covariance model in which most of the eigenspectrum corresponds to noise, while signal is concentrated in a few outlier eigenvalues and eigenvectors. Building on this insight, we denoised these outlier eigenvectors with sparse PCA, selecting the sparsity parameter such that the angle between the inferred subspace and the outlier eigenspace matches the prediction from RMT.

We evaluated our approach on seven datasets spanning seven single-cell RNA-seq technologies and across four sparse PCA implementations. In every case, biwhitening followed by RMT-guided sparse PCA produced subspaces much closer to the signal eigenspace than PCA alone, achieving average noise reduction of $\sim30\%$. These gains translated directly to downstream tasks: on three datasets with ground-truth cell type labels our method reduced $k$-NN classification error compared to PCA and PCA-based approaches, while autoencoder-based methods (scVI, DCA) and the diffusion-based method (MAGIC) failed to improve upon the PCA baseline. Moreover, in terms of $k$-NN classification, sparse PCA approaches achieved performance comparable to that obtained with PCA applied to nearly ten times more cells. Put simply, from the perspective of the PCs, using RMT-guided sparse PCA is equivalent to increasing sample size by an order of magnitude.

The main limitation of our methodology is that, in the absence of a better estimator for the support of $\rho_S$, we are constrained to operate on biwhitened data, for which the support of $\rho_S$ is analytically known. This restriction limits the scope of our approach to providing low-dimensional embeddings with improved signal-to-noise ratio compared to those inferred via standard PCA. Starting from these denoised low-dimensional embeddings, however, it remains unclear how to denoise the raw data itself. While it is tempting, as suggested by our empirical results in Fig.~\ref{fig:fig3} with the unwhitening version of BiPCA, to simply revert the biwhitening procedure, we are not aware of any mathematical guarantee that this will consistently improve the biological signal relative to the raw data. A better estimator for the support of $\rho_S$ would bypass this issue altogether, enabling us to identify, and subsequently correct, the signal eigenspace directly on raw data.

\section{Methods}\label{sec:methods}
\subsection{Eigenspectrum of the sample covariance matrix}
In what follows, we work in the limit $n \to \infty$ with $q = p/n$ fixed. We denote by $S = X^T X / n$ the sample covariance matrix, and by $m(z)$ its Stieltjes transform \cite{yao2015largea}:
\begin{equation}
m(z) = \int \frac{\rho_S(t)}{t - z} dt, \quad z \in \mathbb{C}^+,
\end{equation}
The complementary covariance matrix and its Stieltjes transform are $\underline S = XX^T / n$ and $\underline m(z)$. Random matrix theory (RMT) predicts the limiting spectral density $\rho_S$ as a function of $\rho_A$ and $\rho_B$ \cite{couillet2015analysis, paul2009no, ding2021spiked}. Its Stieltjes transform $m(z)$ is given by \cite{couillet2015analysis, paul2009no}:
\begin{equation}
m(z) = \int \frac{\rho_{B}(t)}{-z(1 + t g_2(z))} dt, \label{sol_stieljes}
\end{equation}
where $g_1(z)$ and $g_2(z)$ are solutions of the self-consistent system:
\begin{align}
g_1(z) &= q \int \frac{t \rho_{B}(t)}{-z(1 + t g_2(z))} dt, \\
g_2(z) &= \int \frac{t \rho_{A}(t)}{-z(1 + t g_1(z))} dt, \quad \forall z \in \mathbb{C}^+. \label{fpoint_g2}
\end{align}
It can be shown that $g_1(z)$ and $g_2(z)$ are also Stieltjes transforms of densities whose supports coincide with the support of $\rho_S$ \cite{couillet2015analysis, ding2021spiked}. These transforms can be extended to $\mathbb{C} \setminus \mathrm{supp} \rho_S$ \cite{couillet2015analysis}. Given $\rho_A$ and $\rho_B$, this system can be solved for $m(z)$, and the density then follows by inversion: $\rho_{S}(x) = \lim_{\eta \rightarrow 0+} \mathrm{Im} m(x + i \eta)/\pi$ \cite{couillet2015analysis}. When $A$ is a low-rank deformation of the identity, we have $\rho_A(t) = \delta(t - 1)$, and Eq.~\ref{fpoint_g2} simplifies with $g_2(z) = \underline{m}(z)$ to yield the Marchenko–Pastur equation \cite{yao2015largea}:
\begin{equation}
m(z) = \int \frac{\rho_{B}(t)}{t(1 - q - q z m(z)) - z} dt.
\end{equation}
Finally, when $\rho_A(t) = \rho_B(t) = \delta(t - 1)$, this equation further simplifies, and the Stieltjes transform $\underline{m}(z)$ has the closed form \cite{yao2015largea}:
\begin{equation}
\underline{m}(z) = \frac{q - 1 - z + \sqrt{(z - 1 - q)^2 - 4q}}{2z}. \label{stieltjes_bw}
\end{equation}
The associated density $\rho_S$ is then the celebrated Marchenko–Pastur distribution.

\subsection{Mapping for left-whitened data}
For left-whitened data $X_{\mathrm{lw}} = A^{-1/2} X$, any outlier eigenvalue $\lambda \notin \operatorname{supp}\rho_S$ corresponds to a signal eigenvalue $\alpha \notin \operatorname{supp}\rho_B$ of $\mathbb{E}[S]$, given by
\begin{equation}
\alpha = -1/\underline m(\lambda),
\end{equation}
where $\underline m(\lambda)$ is the Stieltjes transform of $\underline S$ \cite{yao2015largea}. Beyond eigenvalues, RMT also predicts the squared overlap between an eigenvector $v$ of $S$ associated with $\lambda$ and the low-rank signal subspace of $\mathbb{E}[S]$. Writing $Q$ for the orthogonal projector onto the signal subspace,
\begin{equation}
\|Qv\|^2_2 = \alpha \frac{\psi'(\alpha)}{\psi(\alpha)}, \label{overlap}
\end{equation}
where $\psi$ is the functional inverse of $-1/\underline m(\lambda)$ (see \app{sec:limits}) \cite{yao2015largea}. Reciprocally, for a signal eigenvector $u$ with eigenvalue $\alpha$, the same relation holds for $\|Wu\|^2$, where $W$ is the orthogonal projector onto the outlier eigenspace of $S$ \cite{bloemendal2016principal}. This result also holds with $Q = u u^T$ and $W = v v^T$, provided their eigenvalues are simple and sufficiently separated from the other eigenvalues \cite{bloemendal2016principal, ding2021spiked}. For biwhitened data $\rho_B(t) = \delta (t-1)$ and the Stieltjes transform $\underline m(\lambda)$ is given by Eq.~\ref{stieltjes_bw}, and $\psi$ is given by:
\begin{equation}
\psi(\alpha) = \alpha + q \frac{\alpha}{\alpha - 1},
\end{equation}
which leads directly to Eq.~\ref{criterion_bw}.

\subsection{Datasets and reproducibility}
We use seven publicly available datasets spanning seven single-cell RNA-seq technologies: 10X Chromium v1 (Zheng2017) \cite{zheng2017massively}, Multiome (Luecken2021) \cite{luecken2021sandbox}, inDrops (Hrvatin2018) \cite{hrvatin2018singlecella}, Smart-Seq3xpress (Jensen2022) \cite{hagemann-jensen2022scalable}, Drop-Seq (Macosko2015) \cite{macosko2015highly}, CITE-seq (Stuart2019) \cite{stuart2019comprehensiveb}, and 10X Chromium v3 (Wang2020) \cite{wang2020singlecellb}. For the Multiome dataset, we used only the RNA-seq modality. The dataset were not preprocessed prior to applying the pipelines described in Fig.~\ref{fig:figSpipe}. The only difference across the main text figures is that $2500$ highly variable genes were used for Fig.~\ref{fig:fig1}, while $2000$ highly variable genes were used for all other figures. Complete details about data processing are provided in \app{sec:reproducibility}. The code necessary to reproduce the figures of this paper, as well as python implementations of the Alg.~\ref{algo_sknopp} and Alg.~\ref{fista_spca}, are available in the following github repository: \href{https://github.com/vchz/spcarmt}{https://github.com/vchz/spcarmt}. For both algorithms we use a threshold over the relative improvement on objective functions as stopping criteria.

\section{Acknowledgements}
We thank Michael Shelley, the Biophysical Modeling Group, and the Genomics Group at the Flatiron Institute for valuable discussions. We are also grateful to Giulia Pisegna for feedback on the manuscript. The Flatiron Institute is a division of the Simons Foundation.

%% file: src/si_text.tex
\section{Almost sure limits of outlier eigenvalues in the separable covariance model}\label{sec:limits}
We can rewrite the system of equation for $g_1$ and $g_2$ in the form of a single equation, with $g_2(z)$ the unique solution $g_2$ of $F(z,g_2) = 0$ where $F$ reads
\begin{equation}
F(z, g_2) = g_2 - \int \frac{t \rho_A(t)}{-z\left(1 + tq \displaystyle\int \dfrac{t \rho_B(t)}{-z(1 + t g_2)}dt\right)}  dt.
\end{equation}
With the knowledge of $\rho_A$ and $\rho_B$, assuming we can exactly solve this equation, we have access to $m(z)$, and by the inversion formula to $\rho_S$ and its support.

\subsection{Spiked separable covariance model}
\begin{figure*}[b!]
  \includegraphics[width=\textwidth]{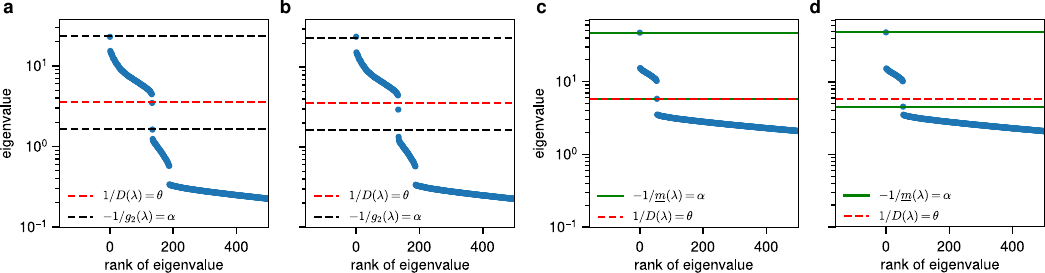}
  \caption{\textbf{Outlier eigenvalues for mixtures of information-plus-noise and separable spiked covariance models}. In this figure we use $n = 3180$ and $p = 3990$. We define a vector $u_1 \in \mathbb{R}^p$ with $u_{1,1} \approx 0.24$, $u_{1,2} \approx 0.97$, and all other entries zero. For the \emph{correlated mixture}, we define a vector $u_2 \in \mathbb{R}^p$ with $u_{2,1} \approx 0.92$, $u_{2,2} \approx 0.39$, and all other entries zero, while for \emph{independent} mixtures $u_2$ has entries chosen uniformly at random. We also define a vector $u_3 \in \mathbb{R}^n$ with entries chosen uniformly at random. All vectors are normalized. The matrices $A$ and $B$ are diagonal. We take $B$ with $60\%$ of its entries equal to $12$ and the rest equal to $1$. The low-rank signals are defined as $Q = 45 u_1 u_1^T$ and $P = 2.5 u_3 u_2^T$. The data is generated as $X = A^{1/2} Y (B + Q)^{1/2} + P$ with entries of $Y$ being independent standard normal variables. The same realization $Y$ is used across all four subfigures. {\bf a, b,} $30\%$ of the entries of $A$ are set to $8$ and the rest to $0.1$. The signal eigenvalues $\alpha$ are computed as the isolated eigenvalues of $(B+Q)$. {\bf c, d}, $A$ is the identity matrix. The signal eigenvalues $\alpha$ are computed as the isolated eigenvalues of $\mathbb{E}[S] = I + Q + P^TP/n$. {\bf a, c.} Independent mixture. {\bf b, d.} Correlated mixture.}
  \label{fig:figSrmt}
\end{figure*}

We consider here that the signal originates in the covariance only, such that $P = 0$. The question we address here is the following: given an eigenvalue $\alpha$ of $B$ such that $\alpha \notin \mathrm{supp} \rho_B$, how is this eigenvalue reflected in the spectrum of $S$? To our knowledge, only partial results for this are available in the literature \cite{ding2021spiked}. The eigenvalue $\alpha$ will give rise to outlier eigenvalues $\lambda \notin \mathrm{supp}\rho_S$ satisfying the following equation
\begin{equation}
g_2(\lambda) = -1/\alpha \label{sep_spike}.
\end{equation}
This equation for $\lambda$ is to be solved outside of $\mathrm{supp} \rho_S$, and if there is no solution then it means that $\lambda$ falls on one of the edges $\rho_S$. We note that this result has only be proven for eigenvalues $\lambda$ above the rightmost edge of the spectrum \cite{ding2021spiked}, but numerical investigation (see Fig.~\ref{fig:figSrmt}) indicate that this result is more general. When there is a solution, there is no guarantee as to whether it is unique, and this equation can have multiple solutions, as also shown in Fig.~\ref{fig:figSrmt}. When $A$ is a low-rank deformation of the identity, i.e. $\rho_A(t) = \delta(t - 1)$, Eq.~\ref{sep_spike} reduces to the classical result for spiked covariance matrices:
\begin{equation}
\underline{m}(\lambda) = -1/\alpha, \label{cov_spike}
\end{equation}
for $\lambda \notin \mathrm{supp}\rho_S$. In this case, because $\underline{m}(z)$ admits a functional inverse outside of $\mathrm{supp} \rho_S$ \cite{yao2015largea}, if there is a solution it is unique, such that one eigenvalue $\alpha$ gives rise to at most one outlier eigenvalue. In the case where $\alpha$ is a spike eigenvalue of $A$ rather than $B$, we can replace $g_2$ by $g_1$ in Eq.~\ref{sep_spike} \cite{ding2021spiked}.

\subsection{Information-plus-noise model with separable covariance}
We consider now the information-plus-noise model with separable covariance. The mean $P > 0$ is a low-rank matrix. To have a non-trivial limit $n \rightarrow \infty$ with $q = p/n$ fixed, the singular values of $P$ need to scale as $\sqrt{n}$. We denote such a singular value $\sqrt{\theta n}$ with $\theta = O(1)$. For this model, the question we address is the following: how is $\theta$ reflected in the spectrum of $S$? Relating $\theta$ to the eigenvalues of $S$ requires the additionnal assumption that the matrices of left and right eigenvectors of $P$ are chosen uniformly at random in the space of orthogonal matrices \cite{benaych-georges2012singular}. In this case $\theta$ gives rise to eigenvalues solution of the equation
\begin{equation}
\lambda \underline{m} (\lambda) m(\lambda) = \frac{1}{\theta}, \label{info_noise}
\end{equation}
with $\lambda \notin \mathrm{supp} \rho_S$ \cite{benaych-georges2012singular, su2025datadriven}. The transform $z \mapsto z m(z) \underline{m}(z)$ is referred as the $D$-transform, and we have no guarantee that a functional inverse exists outside the support of $\rho_S$. For this reason, this equation may have multiple solutions. It was recently shown that in the case where $\rho_A(t) = \delta(t - 1)$, the assumption about the distribution of the eigenvectors of $P$ can be dropped and any eigenvalue $\alpha$ of $A + P^TP/n$ outside the support of $\rho_A$ gives rise to outlier eigenvalues in the spectrum of $S$ that are solutions of equation Eq.~\ref{cov_spike}. This result links the information-plus-noise model with the spiked covariance model: when $\rho_A(t) = \delta(t - 1)$, irrespective of the model, any spike eigenvalue of $\mathbb{E}[S]$ will give rise to a single outlier eigenvalue satisfying criterion \cref{cov_spike}. 

We can rationalize this by verifying that, starting from Eq.~\ref{cov_spike}, we can recover \cref{info_noise} when the matrices of eigenvectors of $P$ are chosen at random in the space of orthogonal matrices. Since $\rho_A(t) = \delta(t - 1)$, the functional inverse of $\underline{m}(z)$ is also explicit \cite{yao2015largea}. Using this inverse, along with $m(z) = (q^{-1} - 1)/z + q^{-1} \underline{m}(z)$, we have 
\begin{equation}
\lambda m(\lambda) \underline{m}(\lambda) = - \int \frac{\rho_B(t)}{t + 1/\underline{m}(\lambda)} dt = -m_B(-1/\underline{m}(\lambda)).
\end{equation}
We recognize $m_B$ the Stieljes transform of $\rho_B$. Because of the assumption on the eigenvectors, we can use a result on low-rank perturbations of symmetric random matrices stating that $\theta$ gives rise to outlier eigenvalues $\alpha$ in $\mathbb{E}[S]$ which are solutions of \cite{benaych-georges2011eigenvalues}:
\begin{equation}
m_{B}(\alpha) =  - 1/\theta \label{theta_random}.
\end{equation}
Combined with the previous equation and \cref{cov_spike}, this allows us to recover \cref{info_noise}. In particular, while \cref{info_noise} can have multiple solutions for $\lambda$, it's now \cref{theta_random} that can have multiple solutions for $\alpha$. In this sense, the result relating outlier eigenvalues of $\mathbb{E}[S]$ to those of $S$ is more general (but not as insightful) than the information-plus-noise equation \cref{info_noise} which requires an additional assumption on the eigenvectors of $P$.

\subsection{Determination of the support}
The previous equations are only usable if we have exact knowledge of the support of $\rho_S$. However, besides the case $\rho_A(t) = \rho_B(t) = \delta (t - 1)$, analytical expressions for the Stieljes transform $m(z)$ and its associated density function $\rho_S$ are not known, and so isn't its support. It was shown that the edges of the support of $\rho_S$, $e_1 > ... > e_K \in \mathbb{R}^+$ of $\rho_S$ can be determined as the real solutions $(x, g_2) = (e_k, g_2(e_k))$ of this system of equation \cite{couillet2015analysis, ding2021spiked}:
\begin{equation}
F(x,g_2) = 0 \text{ and } \frac{\partial F}{\partial g_2}(x,g_2) = 0.
\end{equation}
One can derive a more handy criterion by relating the support to the sign of the derivative $\partial F/\partial g_2$. In particular, it can be shown that any real solutions $(x,g_2)$ to the equation $F(x,g_2) = 0$ with $\partial F/\partial g_2 (x,g_2) > 0$ verifies $x \notin \mathrm{supp} \rho_S$ \cite{couillet2015analysis, ding2021spiked}. This allows us to disregard the condition $\lambda \notin \mathrm{supp} \rho_S$ in all the previous equations, solve them for $\lambda \in \mathbb{R}^{+}$, and check the sign of the derivative $\partial F/\partial g_2 (\lambda,g_2(\lambda)) > 0$. In particular, given a solution $(\lambda, g_2(\lambda))$ to the outlier eigenvalue equation (for the spiked covariance model or the information-plus-noise model), this criterion can be rewritten as:
\begin{equation}
1 - \frac{q}{(\lambda g_2(\lambda) g_1(\lambda))^2} \int \frac{t^2\rho_B(t)}{(t + 1/g_2(\lambda))^2} dt \int \frac{t^2\rho_A(t)}{(t + 1/g_1(\lambda))^2} dt > 0.
\end{equation}
In the case $\rho_A = \delta(t - 1)$, this criterion simplifies, and we recover the one for the spiked covariance model \cite{yao2015largea}, where $\psi$ denotes the functional inverse of $-1/\underline{m}(\lambda)$:
\begin{equation}
\psi'(\alpha) > 0 \text{ with } \psi(\alpha) = \alpha + q \alpha \int \frac{t\rho_B(t)}{\alpha -  t}dt. \label{support_spiked}
\end{equation}
Any outlier eigenvalue associated with a spike eigenvalue $\alpha$ that does not satisfy the criterion will have as almost sure limit a point at an edge of the support of $\rho_S$. For the sake of this paper, it means that it can't be distinguished from the bulk. We note however that by leveraging additional results from RMT this could \emph{in theory} be done \cite{monasson2015estimating}.

\subsection{Choice of a model}
The inverse problem at hand is the following: given outlier eigenvalues observed in $S$, we want to infer the associated signal eigenvalue in $\mathbb{E}[S]$. For this, we need to assume an underlying random matrix model, but we would like to do so with minimal assumptions. We have the following choices:
\begin{enumerate}
    \item Independent mixture of spiked separable covariance and information-plus-noise models: right and left eigenvectors of $P$ are chosen uniformly at random in the space of orthogonal matrices and $\rho_A \neq \delta (t-1)$. In this case, both Eq.~\ref{info_noise} and Eq.~\ref{cov_spike} hold, as shown in Fig.~\ref{fig:figSrmt}a.
    \item Correlated mixture of spiked separable covariance and information-plus-noise models: $P$ and $B$ are not independent and $\rho_A \neq \delta(t-1)$. Eq.~\ref{info_noise} and Eq.~\ref{cov_spike} can't be reliably used to predict the position of outlier eigenvalues, as shown in Fig.~\ref{fig:figSrmt}c.
    \item Correlated or independent mixture of left-whitened spiked covariance and information-plus-noise models: $P$ and $B$ may or may not be independent and $\rho_A = \delta(t-1)$. In this case, Eq.~\ref{cov_spike} relates all outlier eigenvalues of $S$ to those of $\mathbb{E}[S]$, as shown in Fig.~\ref{fig:figSrmt}c,d.
\end{enumerate}
In the first model, we don't have a single mapping relating eigenvalues of $S$ to those of $\mathbb{E}[S]$. Without additional information like fluctuations around the almost sure limits of outliers, we cannot choose between the information-plus-noise or the spiked covariance models. In the second model, to our knowledge we don't have a consistent mapping relating all outlier eigenvalues to signal eigenvalues. Finally, in the third model we do not need any assumption on the left and right eigenvectors of $P$, and we have a unique formula to relate the outlier eigenvalues of $S$ to those of $\mathbb{E}[S]$. To use this model it is necessary to first whiten the cell-cell covariance, i.e. it is only applicable to $X \leftarrow A^{-1/2}X$. 

\section{Reproducibility}\label{sec:reproducibility}
For the sake of reproducibility, we provide flow charts describing the quality control and feature selection steps, as well as the design of the benchmark. These flow charts are shown in Fig.~\ref{fig:figSpipe}, Fig.~\ref{fig:figSpipe_zheng}-Fig.~\ref{fig:figSpipe_stuart}.

\subsection{Quality control and feature selection, Fig.~\ref{fig:figSpipe}}
The default gene selection method is \texttt{flavor='seurat'} from the \texttt{scanpy} package \cite{wolf2018scanpy}. When using \texttt{flavor='seurat\_v3'}, the library-size and log-normalization steps are removed before selecting the set of genes, as recommended by the method. Each pipeline shown in Fig.~\ref{fig:figSpipe} returns count data and serves as the basis for all experiments in this paper.

\subsection{Benchmark design, Fig.~\ref{fig:figSpipe_zheng}-\ref{fig:figSpipe_stuart}}
Each horizontal branch from the main (left-side) pipeline corresponds to an independent copy of the data. Each copy is then processed through the different downstream pipelines. This design ensures that all methods receive exactly the same data, with identical features and cells. The red arrows indicate that the set of genes selected at earlier steps (quality control and highly variable gene filters) is reused across pipelines. This ensures that the full dataset is processed using the same set of genes as the subsampled dataset. Otherwise, the set of genes selected on the full dataset would differ substantially from that obtained on the subsample. These precautions are necessary to guarantee that each method is evaluated fairly and not biased by differences in gene selection.

The final step of each pipeline is a low-dimensional embedding of the cells. For scVI and DCA, each dataset produces two distinct low-dimensional embeddings (PCA and latent) from the same data copy, though the latent embedding is not shown in the flow chart. The parameters displayed correspond to one realization of the trials and the number of components used in the PCA steps may vary slightly. Since DCA and scVI are hyperoptimized, the parameter set for these methods may differ from one trial to another.

%% file: src/si_figures.tex
\clearpage
\newpage
\begin{figure*}
\begin{algorithm}[H]
\caption{FISTA sparse PCA}\label{fista_spca}
\begin{algorithmic}[0]
\State \textbf{input:} sample covariance matrix $S$, leading eigenvectors $V$ (ordered columnwise)
\State \textbf{output:} sparse loading matrix $W$
\State Operations $\mathrm{max}$, $\mathrm{abs}$, $\times$ and $\mathrm{sign}$ are performed entry-wise
\State $p \gets 1/20, \; q \gets 1, \; r \gets 4$
\State $\gamma \gets 1/(2 \lambda_{\max}(S))$, where $\lambda_{\max}(S)$ denotes the max eigenvalue of $S$
\State $W_0 \gets V, \; Y_0 \gets V, \; t_0 \gets 1$
\While{stopping criterion not reached}
    \State $Z \gets Y_{k} + 2\gamma S Y_{k}$
    \State $Z \gets \max\big(\,\mathrm{abs}(Z) - \lambda \gamma, \,0\big) \times \operatorname{sign}(Z)$
    \State $Z \gets \textsc{orthogonalize}(Z)$ \Comment{e.g. Löwdin or Gram-Schmidt}
    \State $t_{k+1} \gets (p + \sqrt{q + r t_k^2})/2$
    \State $Y_{k+1} \gets Z + \dfrac{t_k-1}{t_{k+1}} (Z - W_{k})$
    \State $W_{k+1} \gets Z$
    \State $k \gets k+1$
\EndWhile
\State \Return $W_k$
\end{algorithmic}
\end{algorithm}
\end{figure*}

\clearpage
\newpage
\begin{figure}
  \includegraphics[angle=90, width=0.49\textwidth]{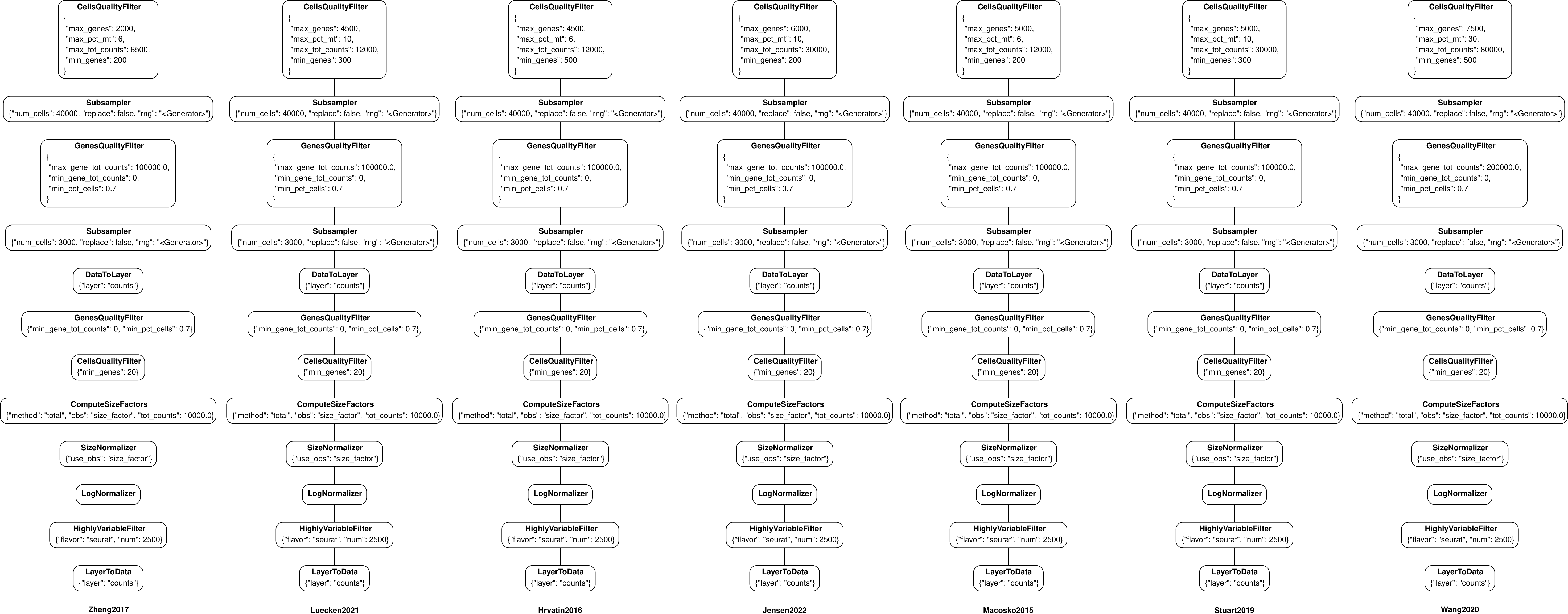}
  \caption{\textbf{Quality control and feature selection pipelines}. This figure is rotated 90 degrees. Each pipeline is read from top to bottom. All quality filters are applied as strict inequalities on the specified thresholds.}
  \label{fig:figSpipe}
\end{figure}
\clearpage
\newpage 
\begin{figure}
  \includegraphics[angle=90, width=0.7\textwidth]{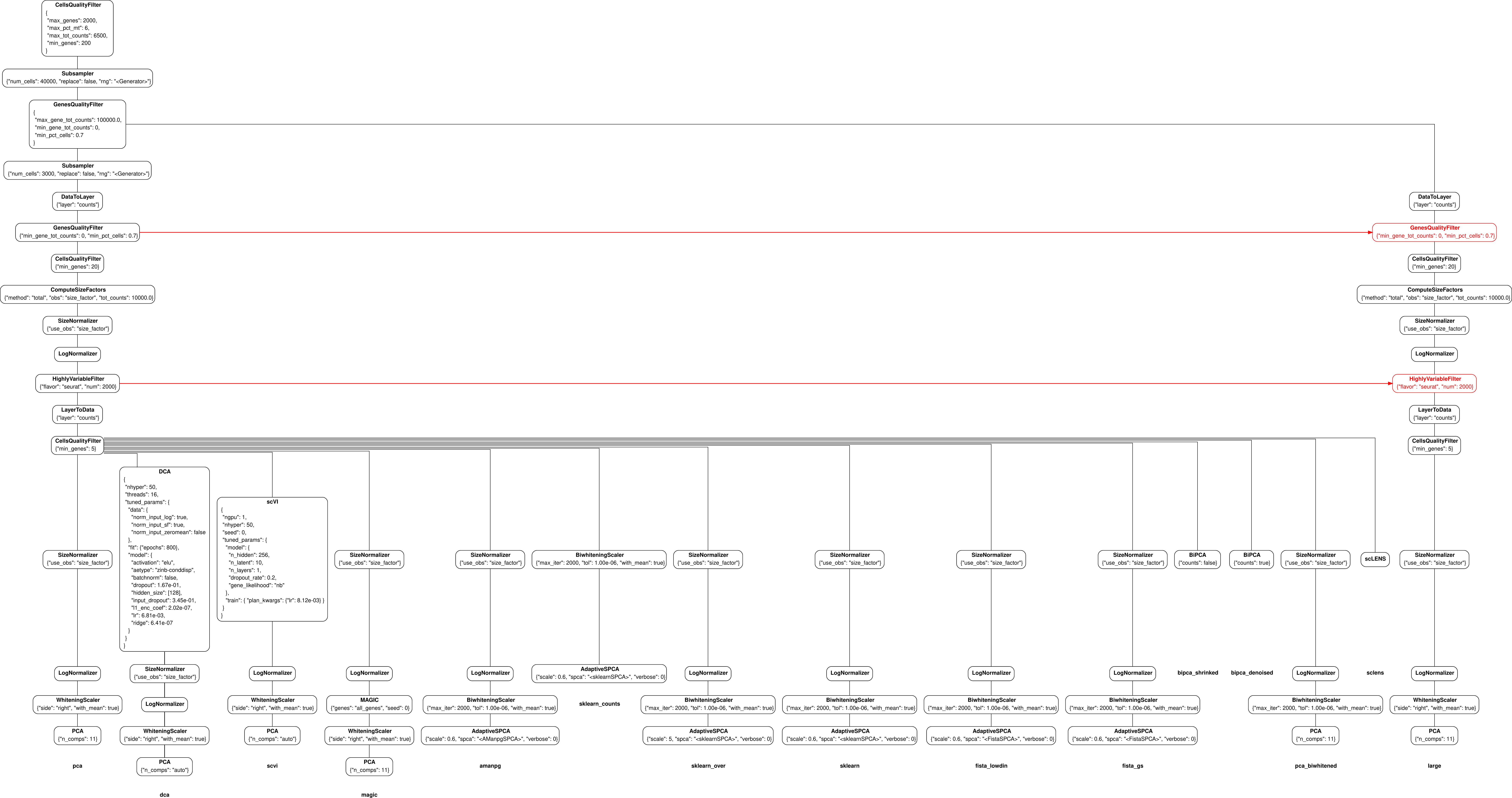}
  \caption{\textbf{Design of the benchmark for Zheng2017}. This figure is rotated 90 degrees. The flow chart is read from top to bottom.}
  \label{fig:figSpipe_zheng}
\end{figure}
\clearpage
\newpage
\begin{figure}
  \includegraphics[angle=90, width=0.7\textwidth]{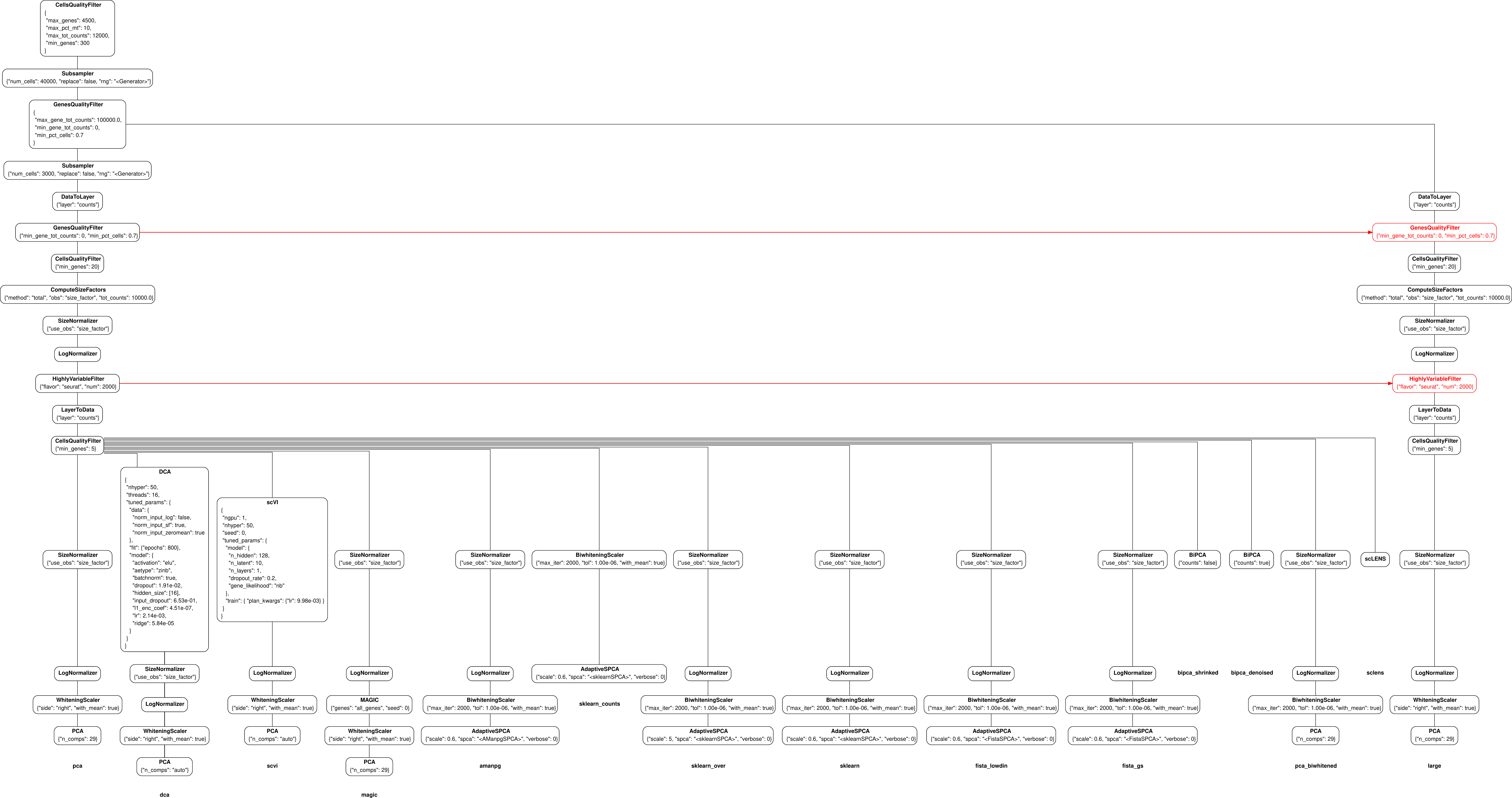}
  \caption{\textbf{Design of the benchmark for Luecken2021}. This figure is rotated 90 degrees. The flow chart is read from top to bottom.}
  \label{fig:figSpipe_luecken}
\end{figure}
\clearpage
\newpage 
\begin{figure}
  \includegraphics[angle=90, width=0.7\textwidth]{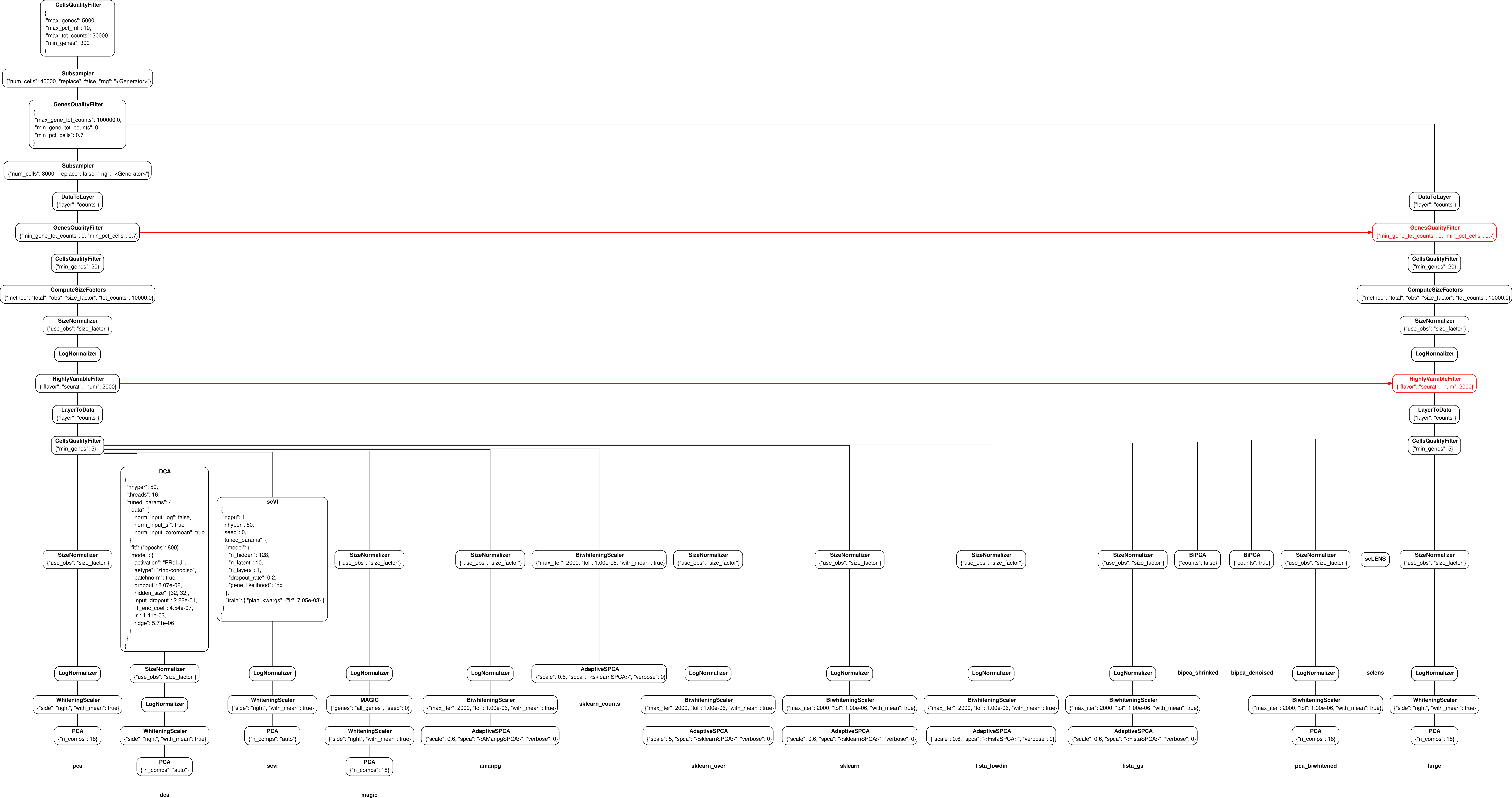}
  \caption{\textbf{Design of the benchmark for Stuart2019}. This figure is rotated 90 degrees. The flow chart is read from top to bottom.}
  \label{fig:figSpipe_stuart}
\end{figure}

\begin{figure}
  \includegraphics[width=\textwidth]{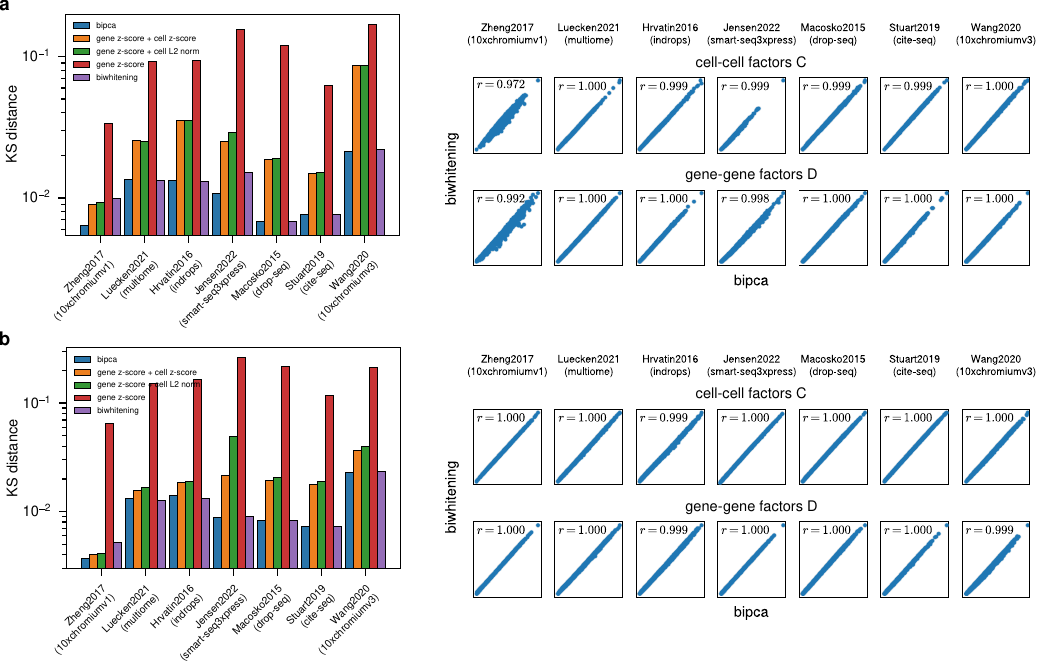}
  \caption{\textbf{Comparison of Biwhitening and BiPCA}. We compare the bi-proportional scaling from the BiPCA package with our biwhitening approach on count data \cite{stanley2025principled} Both methods perform on par, yielding almost identical biwhitening factors. {\bf a.} Results for $2500$ highly variable genes; {\bf b.} results for $10000$ highly variable genes.}
  \label{fig:figScounts}
\end{figure}
\clearpage
\newpage
\begin{figure}
  \includegraphics[width=\textwidth]{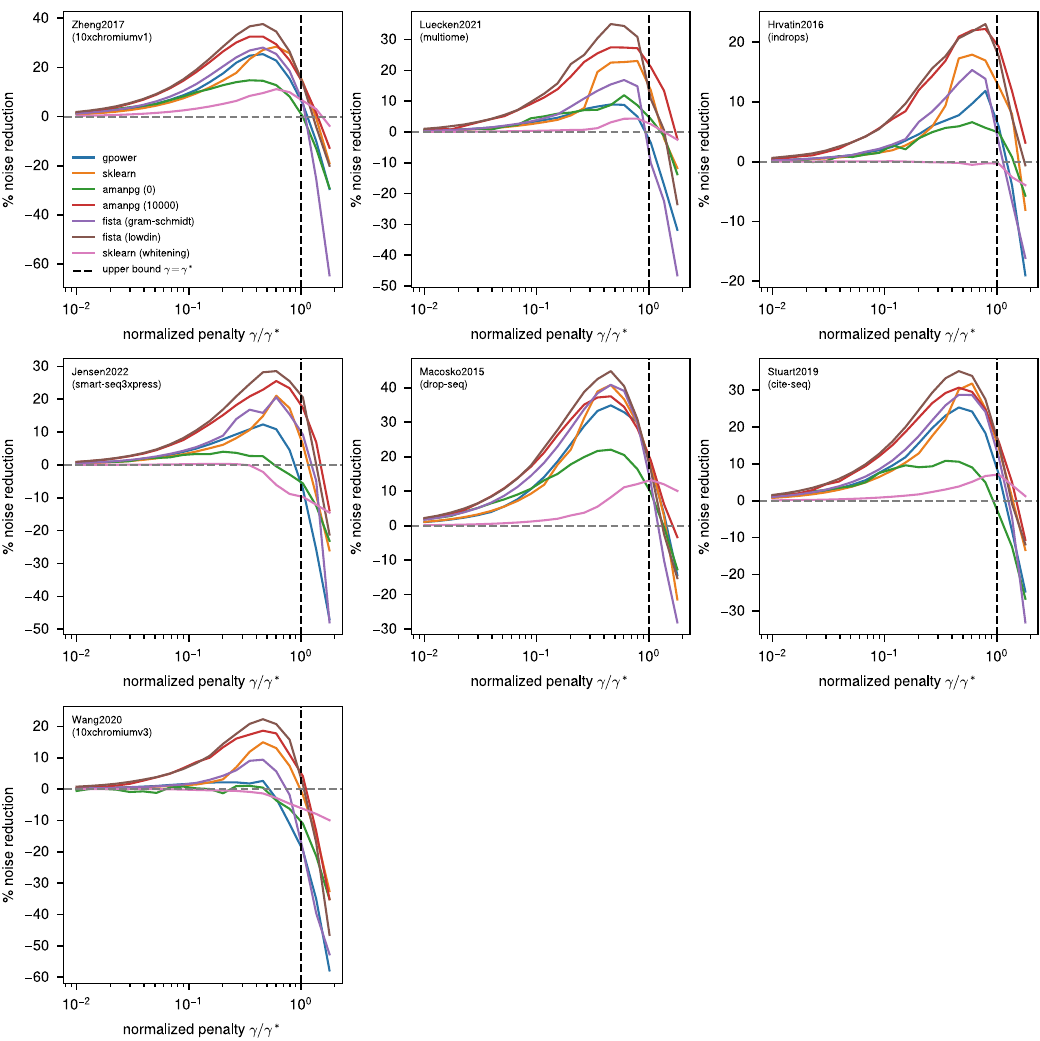}
  \caption{\textbf{Principal subspace reconstruction}. Noise reduction after RMT-guided sparse PCA for all datasets and all sparse PCA algorithm. The results discussed in the main text generalize to all datasets.}
  \label{fig:figSscan}
\end{figure}
\clearpage
\newpage
\begin{figure}
  \includegraphics[width=\textwidth]{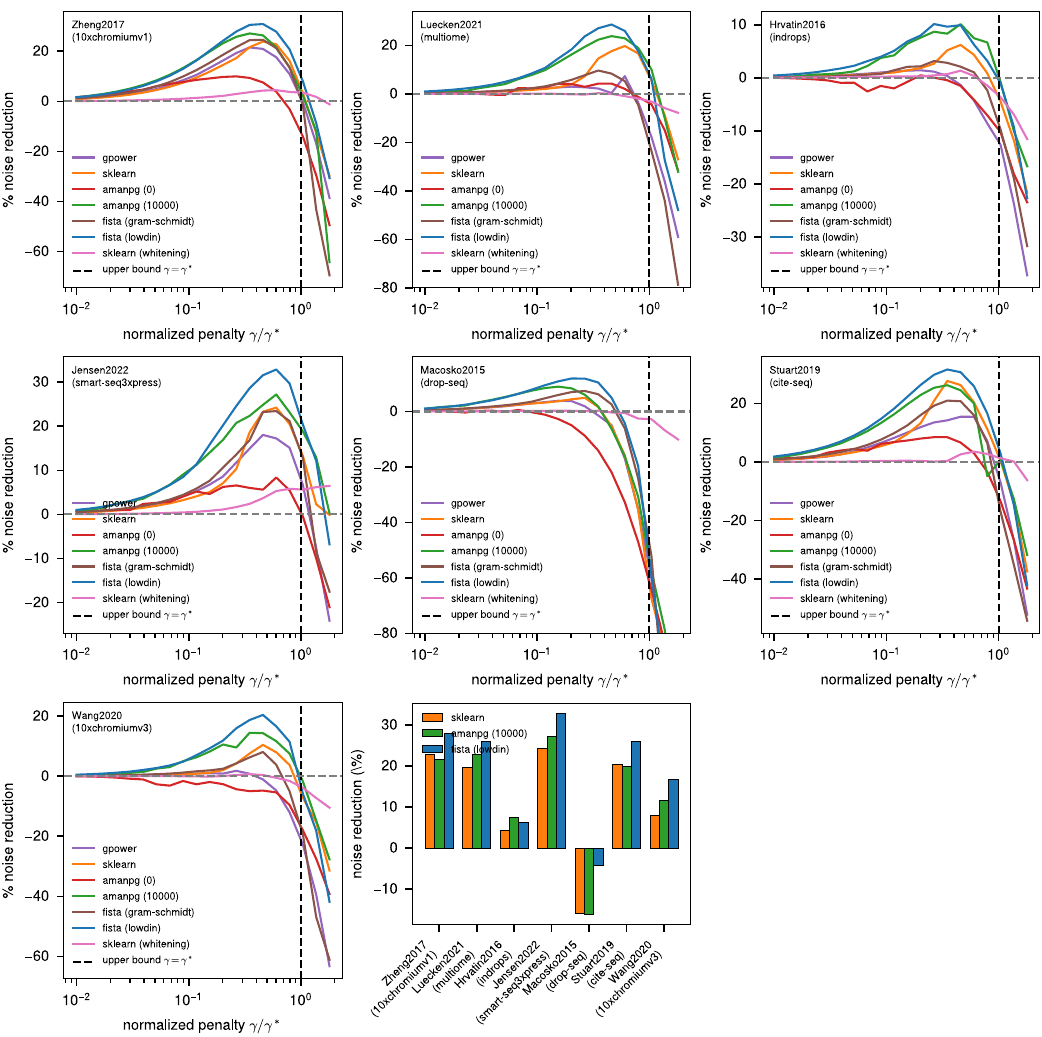}
  \caption{\textbf{Principal subspace reconstruction with different gene sets}. Noise reduction as a function of the penalty parameter for all datasets and all sparse PCA methods using $2000$ highly variable genes selected with the parameter \texttt{flavor='seurat\_v3'} in the \texttt{scanpy} package \cite{wolf2018scanpy}, see Fig.~\ref{fig:figSpipe}.}
  \label{fig:figSscan_v3}
\end{figure}
\clearpage
\newpage 
\clearpage
\newpage 
\begin{figure}
  \includegraphics[width=\textwidth]{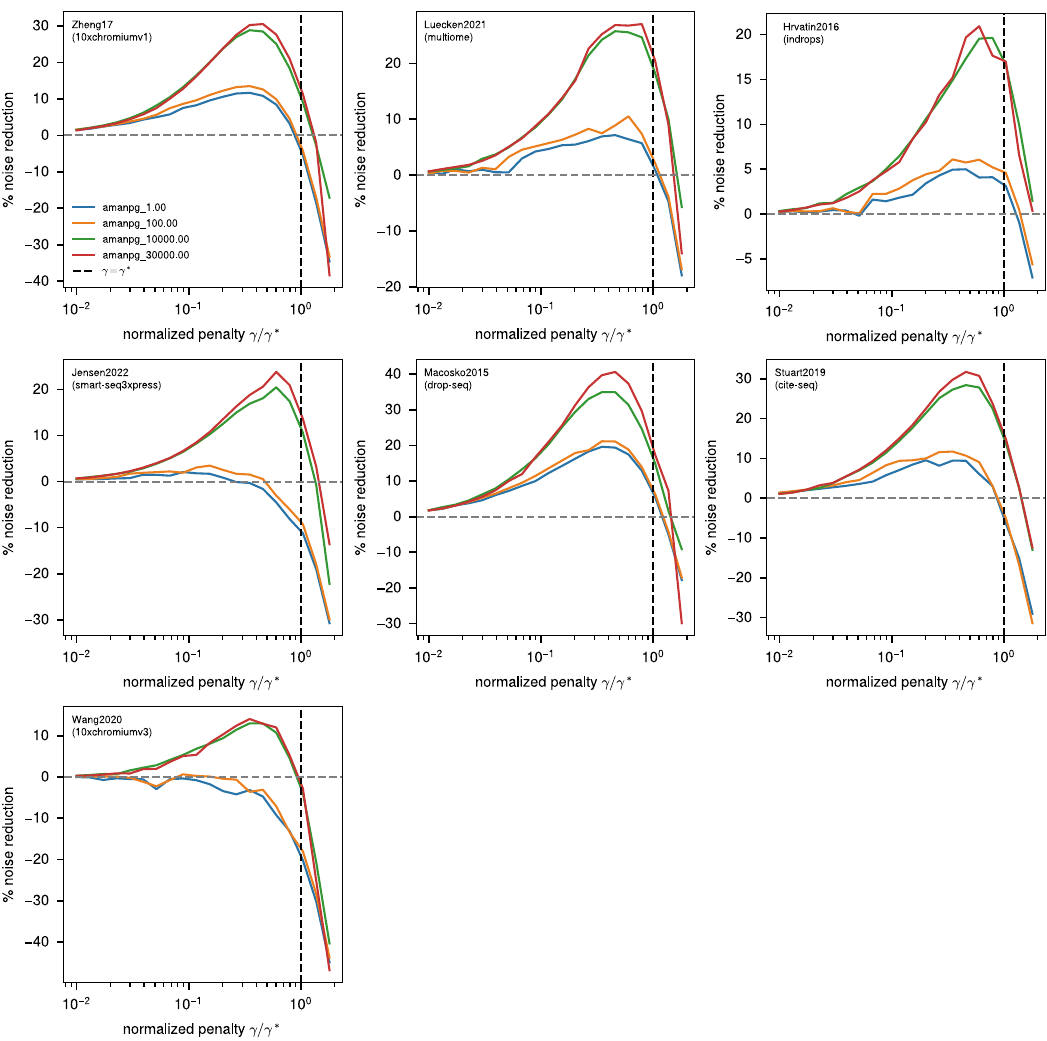}
  \caption{\textbf{AManPG algorithm with different $L_2$ penalties}. Noise reduction as a function of the penalty parameter for all datasets using AManPG with $L_2$ penalties $\eta = 1, 10^2, 10^4, 3 \cdot 10^4$. A very large penalty yields the best performance.}
  \label{fig:figSscan_amanpg}
\end{figure}
\clearpage
\newpage 
\begin{figure}
  \includegraphics[width=0.98\textwidth]{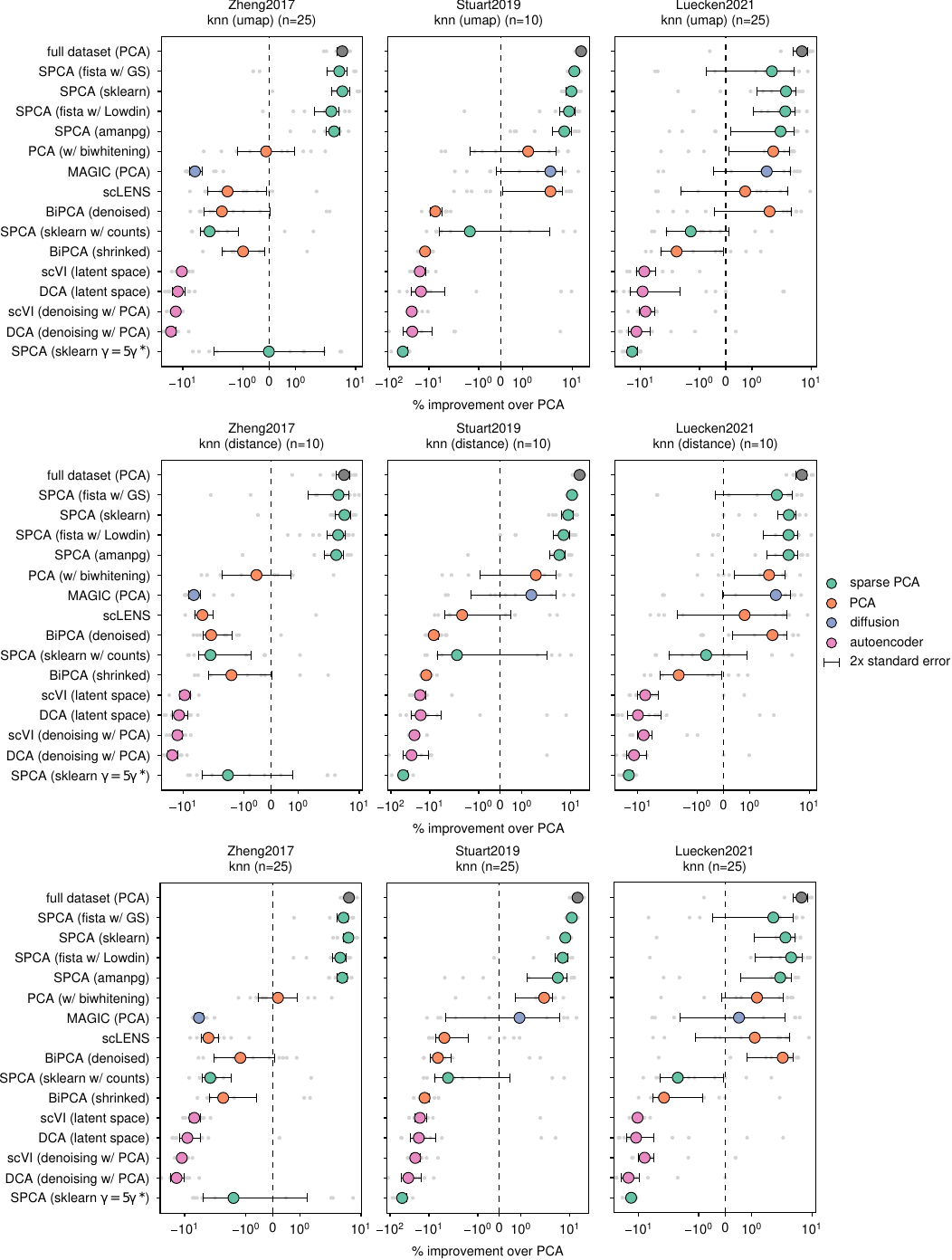}
  \caption{\textbf{$k$-NN classification performance}. Cell type classification performance as evaluated with bagged predictors of $30$ classifiers, measured as a an improvement with respect to the results obtained with PCA. Top row: $n=25$ neighbors with umap weights. Middle row: $n = 10$ neighbors with inverse distance weights. Bottom row: $n=25$ neighbors with constant weights.}
  \label{fig:figSknn}
\end{figure}
\clearpage
\newpage
\begin{figure}
  \includegraphics[width=\textwidth]{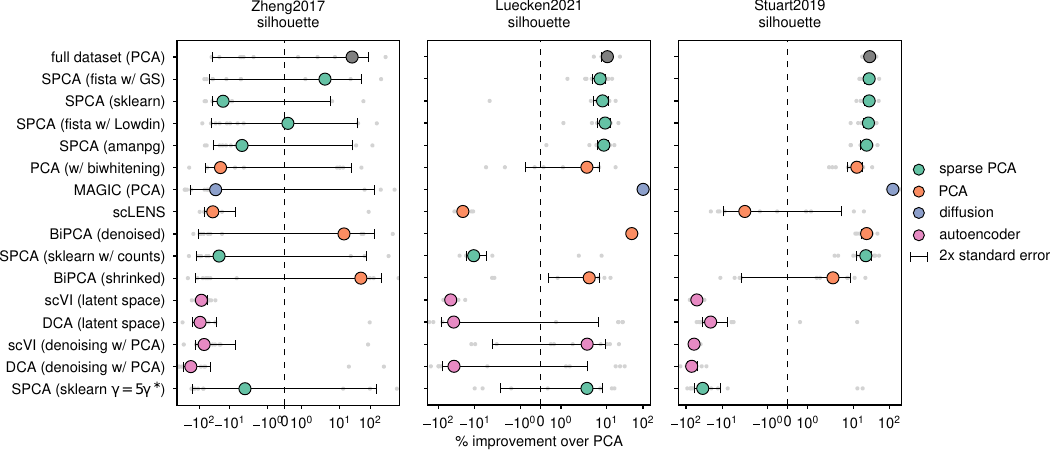}
  \caption{\textbf{Average silhouette score}. Average silhouette score for the ground-truth cell-type annotation in the projected lower-dimensional spaces. The Zheng2017 dataset is particularly challenging, with highly mixed annotations. For this dataset, the average silhouette score of PCA is close to zero, leading to non-significant findings. Both MAGIC and BiPCA perform very well on the two other datasets, despite performing poorly on the $k$-NN benchmark.}
  \label{fig:figSsilhouette}
\end{figure}
\clearpage
\newpage 
\begin{figure}
  \includegraphics[width=0.95\textwidth]{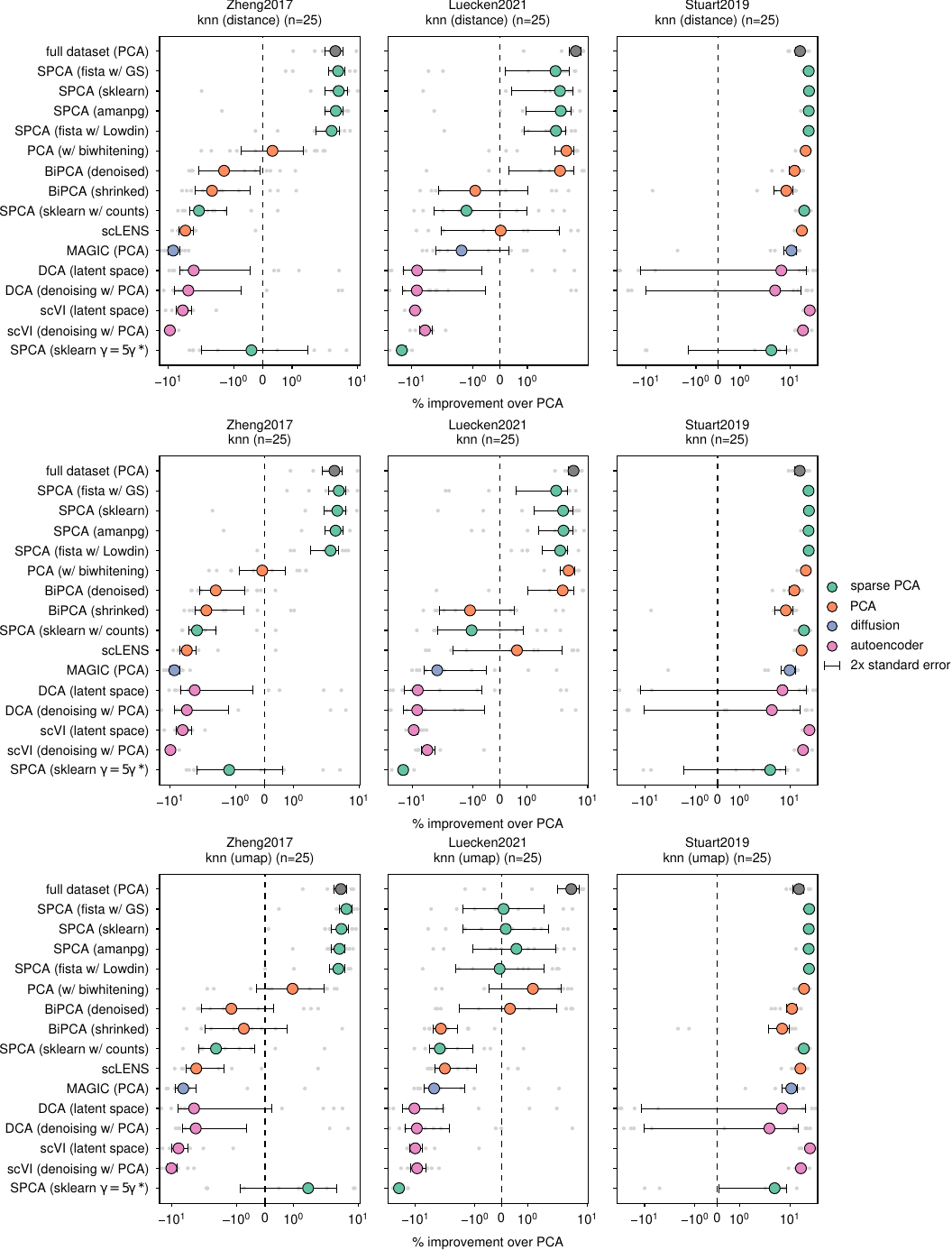}
  \caption{\textbf{$k$-NN classification performance with different gene sets}. Cell type classification performance evaluated using the parameter \texttt{flavor='seurat\_v3'} in the \texttt{scanpy} package \cite{wolf2018scanpy}, see Fig.~\ref{fig:figSpipe}.}
  \label{fig:figSknn_v3}
\end{figure}
\clearpage
\newpage 
\begin{figure}
  \includegraphics[width=\textwidth]{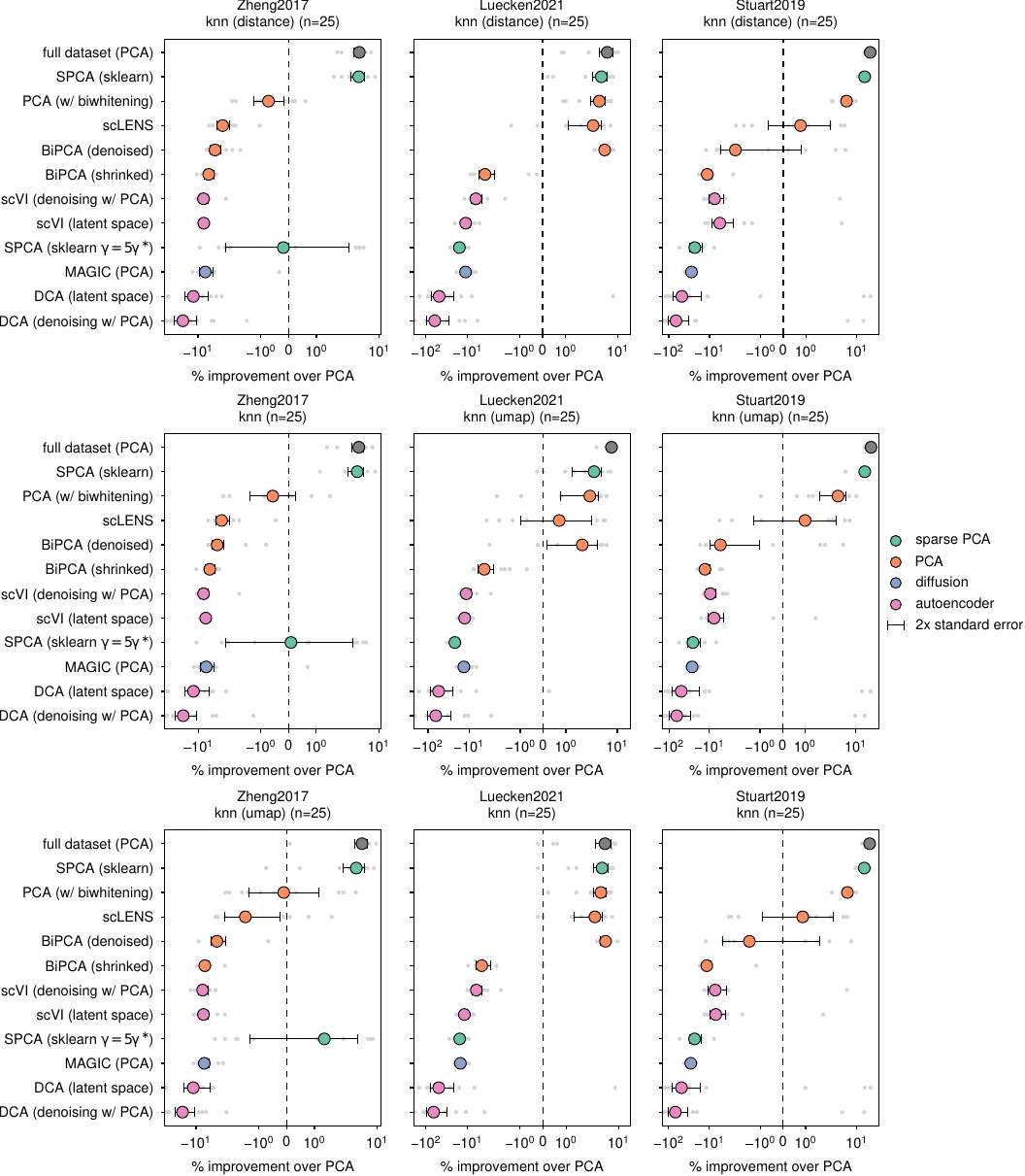}
  \caption{\textbf{$k$-NN classification performance with $10000$ highly variable genes}. Cell type classification performance evaluated Fig.~\ref{fig:figSknn}, but using $n = 10000$ highly variable genes.}
  \label{fig:figSknn_10000}
\end{figure}